\documentclass[twoside]{article}

\usepackage[accepted]{aistats2025}

\usepackage[numbers,compress]{natbib}

\usepackage[utf8]{inputenc} % allow utf-8 input
\usepackage[T1]{fontenc}    % use 8-bit T1 fonts
\usepackage{hyperref}       % hyperlinks
\usepackage{url}            % simple URL typesetting
\usepackage{booktabs, multirow}       % professional-quality tables
\usepackage{amsfonts}       % blackboard math symbols
\usepackage{nicefrac}       % compact symbols for 1/2, etc.
\usepackage{microtype}      % microtypography
\usepackage{xcolor}         % colors
\usepackage{wrapfig} % for side-by
\usepackage{amsthm}
\usepackage{relsize}
\usepackage{mathtools}
\usepackage{graphicx}
\usepackage{pdflscape} 
\usepackage{subcaption}
\usepackage{algpseudocode,algorithm}
\usepackage{paralist}
\usepackage{amsmath, bm}
\usepackage{xcolor}
\usepackage{nccmath}
\usepackage{wrapfig}
\usepackage{tabularx}
\usepackage{upgreek}
\usepackage{pdflscape}% \usepackage[unicode=true,pdfusetitle,
\usepackage{paralist}
\usepackage{subcaption}

\usepackage{hyperref}
\usepackage[nameinlink]{cleveref} %noabbrev
\usepackage{enumitem}
\usepackage{fixfoot}
\usepackage{tikz}
\usepackage{longtable}
\usepackage{mdframed}
\usepackage{tocloft}

\usepackage[breakable]{tcolorbox}

\creflabelformat{equation}{#2\textup{#1}#3}

\usetikzlibrary{bayesnet}

\definecolor{verylightgray}{gray}{0.9} % 95% white and 5% black
\definecolor{veryverylightgray}{gray}{0.97} % 95% white and 5% black

\definecolor{mycol}{rgb}{0,0,0.5}
\hypersetup{
    colorlinks,
    linkcolor={mycol},
    citecolor={mycol},
    urlcolor={mycol}
}

\newtheoremstyle{exampstyle}
  {2.5\topsep} % Space above
  {\topsep} % Space below
  {\itshape} % Body font
  {} % Indent amount
  {\bfseries} % Theorem head font
  {.} % Punctuation after theorem head
  {.5em} % Space after theorem head
  {} % Theorem head spec (can be left empty, meaning `normal')
\newtcolorbox{inferencebox}{
  breakable,
  colback=verylightgray, % Background color of the box
  colframe=black, % Frame color
  coltext=black, % Text color
  boxsep=0pt, % Space between text and the frame of the box
  arc=2mm, % Roundness of the corners
  title=,
  fonttitle=\bfseries,
  left=1mm,
  right=1mm,
  top=1mm,
  bottom=1mm,
  boxrule=0.1mm
}

\theoremstyle{exampstyle}
\theoremstyle{exampstyle}
\theoremstyle{exampstyle}
\theoremstyle{exampstyle}
\theoremstyle{exampstyle}
\theoremstyle{exampstyle}
\theoremstyle{exampstyle}
\theoremstyle{exampstyle}
\newtheorem*{thm*}{Theorem}
\newtheorem*{claim*}{Claim}

\crefname{thm}{theorem}{theorems}
\Crefname{thm}{Theorem}{Theorems}
\crefname{lem}{lemma}{lemmas}
\Crefname{lem}{Lemma}{Lemmas}
\crefname{cor}{corollary}{corollaries}
\Crefname{cor}{Corollary}{Corollaries}
\crefname{defn}{definition}{definitions}
\Crefname{defn}{Definition}{Definitions}
\crefname{assump}{assumption}{assumptions}
\Crefname{assump}{Assumption}{Assumptions}

\global\def\E{\operatornamewithlimits{\mathbb{E}}}%

\global\def\pars#1{\left(#1\right)}%

\global\def\KL#1{[#1]}%

\global\def\KL#1#2{\textrm{KL}\pars{#1\ \middle\Vert\ #2}}%

\newcommand{\margwass}{marginal-Wasserstein}
\newcommand{\Margwass}{Marginal-Wasserstein}

\newcounter{relctr} %% <- counter for relations
\definecolor{mypurple}{rgb}{0.5019607843137255, 0.0, 0.5019607843137255}

\AtBeginDocument{} %% <- store original definition

\allowdisplaybreaks

\newtcolorbox{examplebox}[1][]{%
    colback=veryverylightgray,
    colframe=verylightgray,
    coltitle=black,
    title=#1,
    boxrule=0.5pt,
    boxsep=2pt,
    left=5pt,
    right=5pt,
    top=5pt,
    bottom=5pt,
    fonttitle=\bfseries,
    breakable,
}

\begin{document}

\begingroup
\renewcommand{\addcontentsline}[3]{}

\twocolumn[

\aistatstitle{Disentangling impact of capacity, objective, batchsize, estimators, and step-size on flow VI}

\aistatsauthor{  Abhinav Agrawal \And  Justin Domke} 
\aistatsaddress{University of Massachusetts Amherst \And University of Massachusetts Amherst } 
]
\begin{abstract}
  Normalizing flow-based variational inference (flow VI) is a promising approximate inference approach, but its performance remains inconsistent across studies.
  Numerous algorithmic choices influence flow VI's performance.
  We conduct a step-by-step analysis to disentangle the impact of some of the key factors: capacity, objectives, gradient estimators, number of gradient estimates (batchsize), and step-sizes.
  Each step examines one factor while neutralizing others using insights from the previous steps and/or using extensive parallel computation.
  To facilitate high-fidelity evaluation, we curate a benchmark of synthetic targets that represent common posterior pathologies and allow for exact sampling. 
  We provide specific recommendations for different factors and propose a flow VI recipe that matches or surpasses leading turnkey Hamiltonian Monte Carlo (HMC) methods. 
\end{abstract}

\section{INTRODUCTION}

\begin{figure*}
  \centering
  \includegraphics[trim = 0 0 0 30, clip, width=0.85\linewidth]{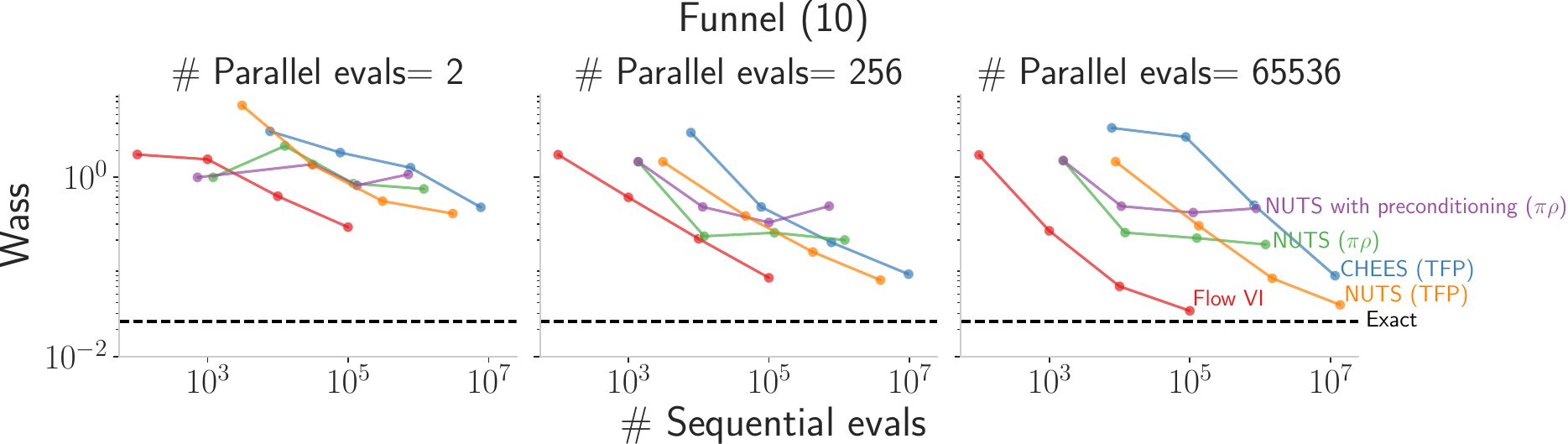}
  \caption{\small{{\Margwass} metric (\cref{eq: wass}) against sequential evaluations for Neal's funnel \citep{neal2001annealed} in ten dimensions, with parallel budget increasing from left to right. 
  For flow VI, sequential evaluations count optimization iterations and parallel evaluations represent batchsize. 
  For HMC, sequential evaluations count leapfrog steps and parallel evaluations represent number of chains.  
  (We use implementations from NumPyro and TensforFlow Probability, denoting these with $\pi\rho$ (read ``pyro'') and TFP, respectively.) 
  The black dotted line indicates the {\margwass} metric under exact samples (\Cref{sec:eval}). 
  Flow VI is almost as accurate as exact inference and faster than HMC when using sufficient parallel budget.
}}
  \label{fig:hmc_vs_flow_vi}
  \vspace{0pt}
\end{figure*}

Normalizing flow-based variational inference (flow VI) \citep{rezende2015variational} approximates posterior distributions \citep{webb2019improving,aagrawal2020,ambrogioni2021automatic_b,glockler2022variational,durr2024bernstein} by constructing flexible variational families through a series of invertible transformations \citep{papamakarios2019normalizing,kobyzev2020normalizing}. While promising, performance of flow VI remains inconsistent\textemdash some studies report success \citep{aagrawal2020,ambrogioni2021automatic_b,glockler2022variational,vaitl2022gradients,durr2024bernstein}, while others highlight optimization challenges and poor outcomes \citep{hoffman2020black,behrmann2021understanding,baudart2021automatic,dhaka2021challenges,jaini2020tails,liang2022fat,andrade2024stable}. 

Multiple factors influence flow VI, making it challenging to determine the cause of inconsistencies: 
\emph{Is it insufficient capacity, inappropriate objective, high gradient variance, or incorrect step-size?}
The "entangling" of such factors limits understanding of flow VI's capabilities and best practices, hindering its wider adoption.

This paper takes a step-by-step approach to disentangle the impact of key factors\textemdash capacity (\Cref{sec:capacity}), objective (\Cref{sec:divergences}), gradient estimator and batchsize (\Cref{sec:estimators}), and step-size (\Cref{sec:optimization}). 
Each step answers a fundamental question about a factor while neutralizing the impact of others by either applying insights from the previous steps or using compute of modern GPU clusters. For example, \Cref{sec:divergences} asks:\emph{ Do we need complicated mode-spanning objectives?} To test this, we neutralize the impact of capacity by borrowing tested flow architectures from  \Cref{sec:capacity} and neutralize the impact of optimization choices by using a massive batchsize alongside extensive hyperparameter sweeps. 

This approach requires precise measures of accuracy.
However, intractable posteriors complicate high-fidelity evaluations as without true samples, common metrics (applicable to both VI and HMC) are rare\textemdash the evidence lower bound and its variants \citep{burda2015importance,dieng2017variational,domke2018importance,domke2019divide} work only for VI, while convergence measures like effective sample size are used primarily for HMC. Although one might use samples from an established method as proxies for ground truth \citep{magnusson2024posteriordb}, reliable, common metrics are still limited\textemdash the Wasserstein distance \citep{villani2009optimal} is often used but scales poorly with sample size \citep{cuturi2013sinkhorn}, making it unsuitable for high-fidelity evaluations (\Cref{sec:wass}).

To overcome evaluation issues, we curate a benchmark of synthetic targets (\Cref{sec:models}) that reflect common posterior pathologies: poor conditioning, nonlinear curvature, heavy tails, and parameter interdependence. 
These targets provide ground truth samples, ensuring high-fidelity evaluations (see \Cref{sec:eval} for evaluation strategy). 
Additionally, we use a scalable proxy for the Wasserstein distance\textemdash{\margwass} (\cref{eq: wass})\textemdash which averages the Wasserstein distances between one-dimensional marginals. This scales log-linearly in sample size, offering a precise and efficient comparison of VI and HMC methods (\Cref{sec:wass}).

We present our major findings below.

% \begin{itemize}[leftmargin=15pt,itemsep=0pt,topsep=0pt]

  % \item 
  \textbf{Capacity.} Real-NVP flows \citep{dinh2016density} can accurately approximate challenging targets (\Cref{fig: capacity}). 
  While commonly used \citep{webb2019improving,aagrawal2020,glockler2022variational,xu2023mixflows}, some works report poor outcomes with real-NVP \citep{behrmann2021understanding,dhaka2021challenges,jaini2020tails,andrade2024stable}. 
  We neutralize the impact of other factors and show that, with appropriate architectural choices, real-NVP has sufficient capacity to represent challenging targets accurately (\Cref{sec:capacity}). 

  % \item 
  \textbf{Objectives.} With high-capacity flows, complicated objectives are unnecessary (\Cref{fig: objective wass}). 
  Some works advocate for mode-spanning objectives due to their better coverage \citep{li2016renyi,wang2018variational,naesseth2020markovian}, but others show that these are hard to optimize \citep{geffner2020difficulty,geffner2021empirical}.
  We directly optimize mode-spanning $\KL{p}{q}$ using exact samples and show this indeed achieves great results (\Cref{fig: objective wass}). 
  However, the easier to optimize mode-seeking $\KL{q}{p}$ is sufficient when capacity is high (\Cref{fig: objective wass}).

  % \item 
  \textbf{Gradient batchsize and estimators.} Large batchsize greatly benefit high-capacity flows (\Cref{fig: estimators}). 
  Some works explore lower-variance gradient estimators for flows \citep{aagrawal2020,vaitl2022gradients,vaitl2024fast}. While these help, we show they are insufficient alone, and a larger batchsize significantly improves the performance of high-capacity flows, suggesting combined usage when possible (\Cref{sec:estimators}).
  
  % \item 
  \textbf{Step-size and optimization.} Maintaining the step-size within a narrow range is crucial for convergence (\Cref{fig: optimization}).
  Automating the choice of step-size is an open problem \citep{kucukelbir2017automatic,welandawe2024}. We show that even with high-capacity flows, low gradient variance, and adaptive optimizers \citep{kingma2014adam}, optimization can diverge abruptly after appearing to work for thousands of iterations (\Cref{sec:optimization}). Step-sizes within $10^{-4}$ to $10^{-3}$ provide consistent results across targets over long runs (\Cref{fig: optimization}).

% \end{itemize}

Overall, two choices are essential: high-capacity flows (solves representational constraints) and large gradient batchsizes (simplifies optimization). 
With these, a simple recipe suffices: use the standard VI objective, reduced variance gradient estimator (if possible), and a fixed step size within a small range; appropriately initialize and parameterize flow architectures, and optimize for long using an adaptive optimizer like Adam \citep{kingma2014adam} (see \Cref{app: sec: recipe} for details). 
Using this recipe, we show flow VI matches or surpasses leading HMC methods on challenging targets in fewer model evaluations (see \Cref{sec:batch_size} and \Cref{fig: batchsize and hmc,fig:hmc_vs_flow_vi}).

\section{SETUP}
\label{sec:problem_setup}

\begin{figure*}[ht!]
  \centering
    \includegraphics[width =0.85\textwidth]{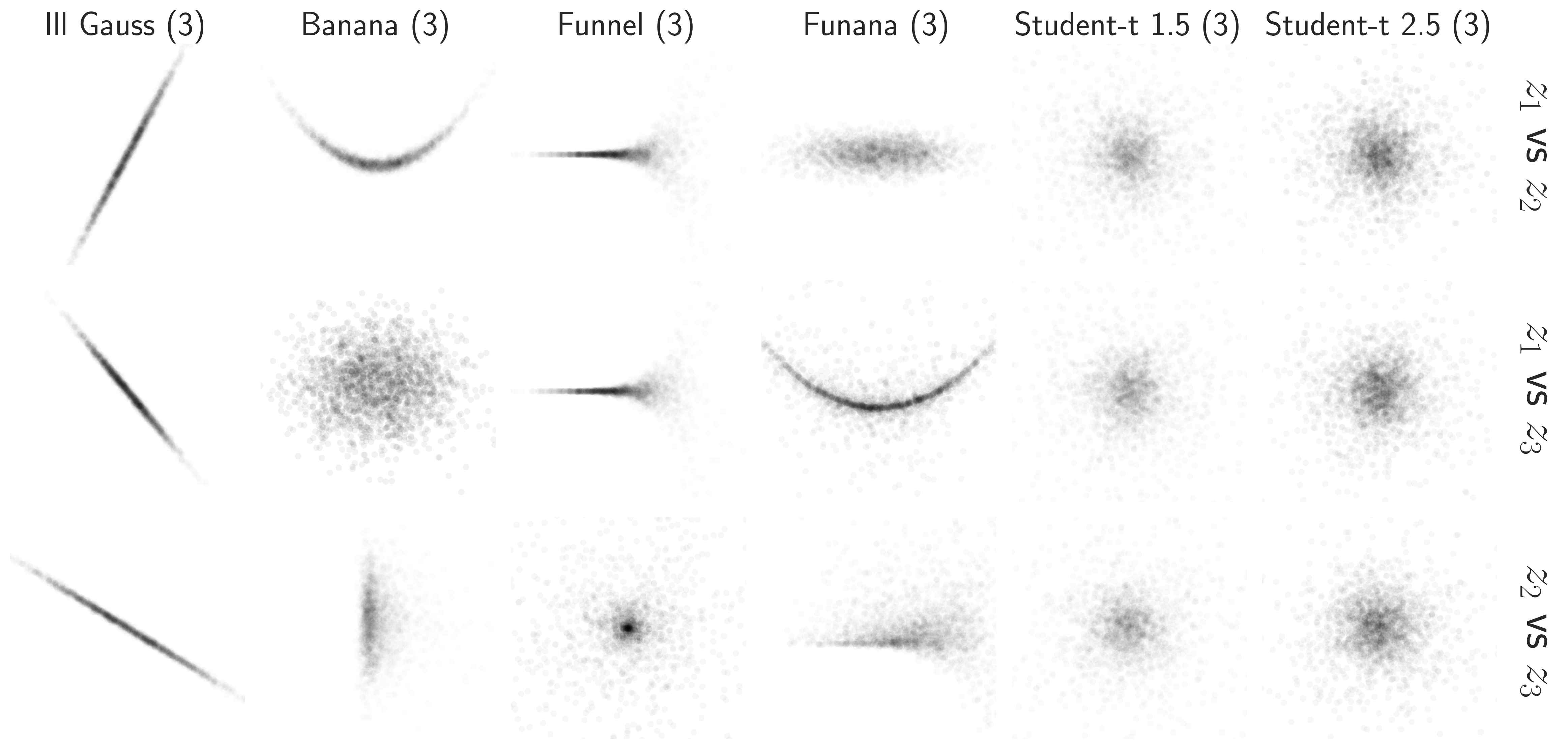}
  \caption{\small{\label{fig: models} \textbf{Rows:} Pair marginals. These targets cover various pathologies: Ill-conditioned Gaussian has high correlations, Banana has non-linear relationships, Neal's funnel has parameters whose spread depends on other parameters, Funana combines funnel-like behavior with non-linearity, and Student-t with $\nu = 1.5$ has heavier tails than Student-t with $\nu = 2.5$.}}
  \vspace{0pt}
\end{figure*}

  Given a model $p(z, y)$ where $z$ are the latent variables and $y$ are the observed variables, the goal of inference is to approximate the posterior $p(z \vert y).$ 
  Variational inference learns an approximation $q$  by maximizing evidence lower-bound (ELBO)
  \citep{saul1996mean,jordan1999introduction,wainwright2008graphical,blei2017variational}, where
 \begin{align}
   \textrm{ELBO} &\coloneqq \E_{q} [ \log p(z, y) - \log q(z) ]. \label{eq:ELBO}
\end{align}
We focus on classical probabilistic models where  $p(z,y)$ does not have any unknown parameters apart from $z$ (different from generative models \citep{kingma2013auto,rezende2014stochastic}). Maximizing the ELBO is equivalent to minimizing $\KL{q}{p}$. %Once the optimized, sampling from $q$ approximates sampling from $p(z \vert y)$.
  
  % \subsection{Variational Family}
  % \label{sec:var fam}
  We use normalizing flows \citep{rezende2015variational,kobyzev2020normalizing,papamakarios2019normalizing} as the variational family for $q$. The main idea behind flows is to transform a base density $q_\epsilon(\epsilon)$ using a diffeomorphism $T$. Let $\epsilon \sim q_\epsilon(\epsilon)$. Then, the transformed variable $z = T(\epsilon)$ has the density
  $q(z) = q_\epsilon(\epsilon) |\det \nabla T(\epsilon)|^{-1}.$ Usually,  $T$ is composed of a sequence of neural-network-based transformations designed such that $T$ is invertible and the determinant of the Jacobian $|\det \nabla T(\epsilon)|$ can be calculated efficiently \citep{dinh2014nice,dinh2016density}.
  
  \label{sec: var fam}
  We use real-NVP flows \citep{dinh2016density} due to their simplicity and popularity \citep{webb2019improving,aagrawal2020,thin2020metflow,dhaka2021challenges,samsonov2022local,glockler2022variational,siahkoohi2023reliable,xu2023mixflows,andrade2024stable}. 
  These flows are composed of a sequence of layers, where each layer applies an affine transformation to one-half the variables, with the scale and translation parameters of the affine transform given by a neural network that takes the other half of the variables as input (see \Cref{app:sec:details of real-NVP} for details).
  These flows have an efficient forward ($T$) and inverse ($T^{-1}$) pass, allowing additional gradient estimators \citep{vaitl2022gradients,vaitl2024fast} to be used in \Cref{sec:estimators}.

  \subsection{Targets}
  \label{sec:models}
  % \jd{hard to tell what the references to section 2.2 vs. 3 and 4 mean here}
  We design a benchmark of synthetic targets, capturing common pathologies in real-world posteriors. (See \Cref{app: sec: details on other models} for target details.)
  All of these targets allow exact sampling, enabling analysis that is otherwise impossible (\Cref{sec:capacity,sec:divergences}). 
  The exact samples also allow comparisons for fair evaluations of VI and HMC methods (see \Cref{sec:eval} for evaluation strategy).
  See \Cref{fig: models} for samples from these targets in three dimensions.

\begin{figure*}[ht!]
  \centering
  \includegraphics[trim = 0 0 0 0, clip, width = 0.85\textwidth]{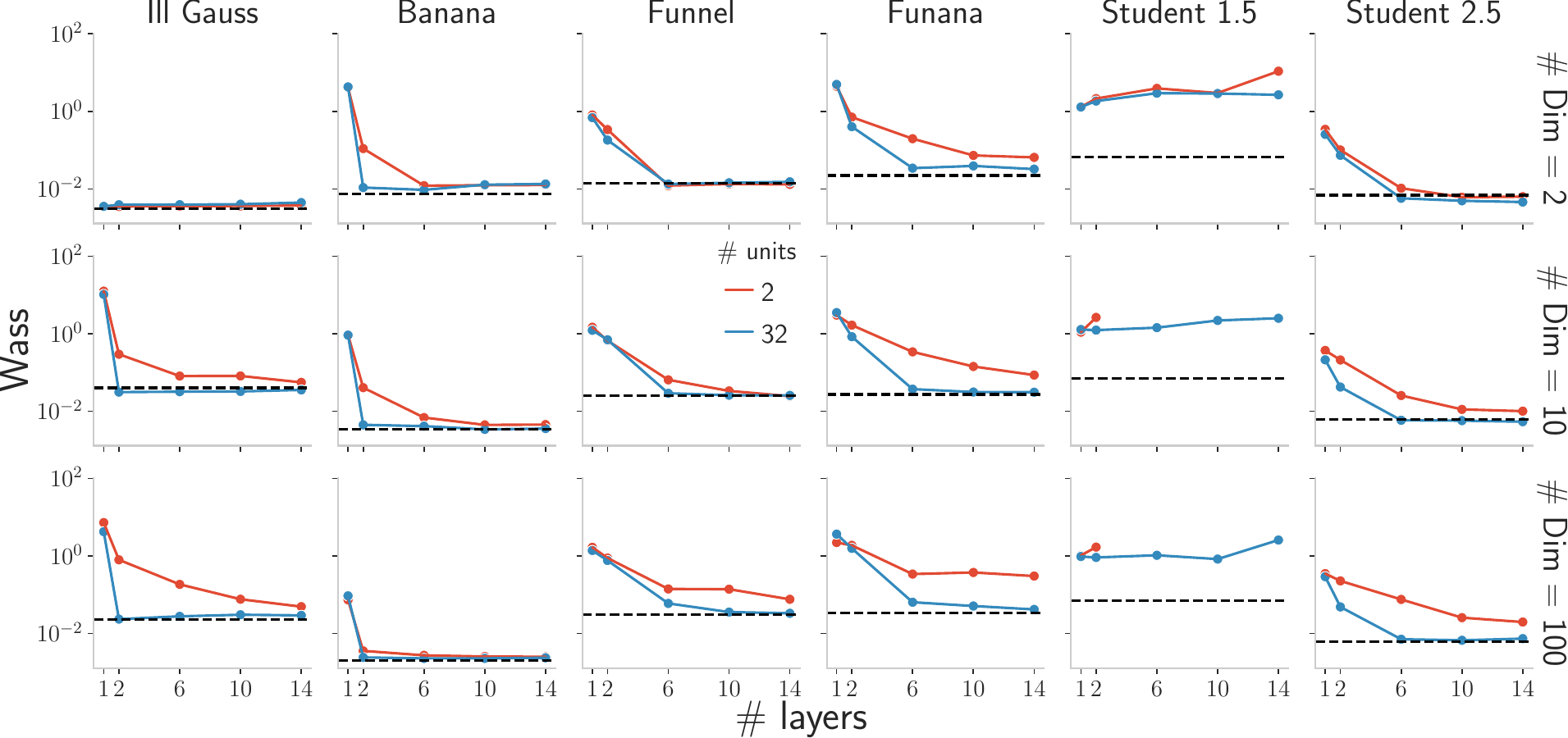}
  \caption{\small{\label{fig: capacity} \textbf{Rows:} Model dimensions. {\Margwass} metric (\cref{eq: wass}) against number of coupling layers for different number of hidden units. Performance improves with increase in either of these levers of capacity for all targets but Student-t with  $\nu = 1.5$. Heavy tails of this target create problems when optimizing $\KL{p}{q}$ \citep{jaini2020tails}. See \Cref{fig: objective wass} for comparisons with $\KL{q}{p}$ optimization. (In the first row, Funana uses 3 dimensions.) } }
  \vspace{0pt}
\end{figure*}

  \subsection{Metrics}
  \label{sec:wass}

  Several metrics of performance apply to either  VI or HMC, but not both: 
  standard metrics like ELBO, evidence upper bound \citep{ji2019stochastic}, and importance-weighted ELBO \citep{burda2015importance} do not apply to HMC;
  convergence diagnostics like effective sample size do not apply to VI.
  Other metrics make precise comparisons hard: predictive test likelihoods are unreliable due to the vagaries of test datasets \citep{agrawal2024understanding} and moments might not exist for heavy-tailed targets like Student-t with $v = 1.5$. 
  
  The Wasserstein distance \citep{villani2009optimal} is rare in that it allows fair VI and HMC comparisons, but requires access to reference samples and scales poorly in sample-size \citep{cuturi2013sinkhorn}. In preliminary experiments, we found the noise in the Wasserstein distance estimates made fine comparisons impossible, even when using thousands of samples (see \Cref{fig: wasserstein empirical demonstration} in \Cref{app: sec: details of wasserstein evaluation.}). 
  
  Fortunately, calculating the Wasserstein distance in the univariate case reduces to sorting.
  Based on this, we use {\margwass} metric\textemdash the average Wasserstein distance between the corresponding one-dimensional marginal distributions.
  Let both $A$ and $B$ be $S \times d$ sample matrices, where $S$ is the number of samples, and $d$ is the number of dimensions. Let $A^{\#}$ be the column-wise sorted version of $A$ such that $A^{\#}_{0j}$ is the smallest value in column $j$. Then, 
  \begin{align}
      &\text{\margwass} (A, B) \coloneqq  \nonumber\\
      & \phantom{some text will help so}  \mathsmaller{ \frac{1}{d} \sum_{j=1}^{d}  \frac{1}{S} \sum_{i=1}^{S}  \vert A^{\#}_{ij} - B^{\#}_{ij} \vert}. \label{eq: wass}
      % \text{\margwass} (A, B) \coloneqq  \customsmaller[0.8]{ \frac{1}{d} \sum_{j=1}^{d}  \frac{1}{S} \sum_{i=1}^{S}  \vert A^{\#}_{ij} - B^{\#}_{ij} \vert}. \label{eq: wass}
      % \text{\margwass} (A, B) \coloneqq  { \frac{1}{d}  \frac{1}{S} \sum_{i=1,j=1}^{d,S}  \vert A^{\#}_{ij} - B^{\#}_{ij} \vert}. \label{eq: wass}
      % \text{\margwass} (A, B) \coloneqq  \frac{\sum_{j=1}^{d} \sum_{i=1}^{S} \vert A^{\#}_{ij} - B^{\#}_{ij} \vert}{d S}. \label{eq: wass}
    \end{align}
Like the Wasserstein distance, this also suffers from Monte Carlo noise when the sample-size is relatively small, but since it is more scalable, we can use a much larger sample-size to reduce the noise (see \Cref{app: sec: details of wasserstein evaluation.}). 
Of course, the {\margwass} metric only looks at marginals, and so it misses the correlations between dimensions. However, its scalability makes it the only reliable choice for high-fidelity evaluations.
% \jd{Need to acknowledge that only looks at marginals, could be "missing" something in general}
  % (We use one million samples for our evaluations). 

  \subsection{Evaluation Strategy}
  \label{sec:eval}  
  For reliable performance evaluations, we use marginal-Wasserstein metric (\cref{eq: wass}). 
  %Theoretically, when an inference is exact, the Wasserstein distance is zero \citep{villani2009optimal}. 
  %However, in practice, this value depends on the sample-size and the target geometry \citep{cuturi2013sinkhorn}. 
  The minimum achievable value of this metric, even with exact inference, depends on the sample size and target geometry \citep{cuturi2013sinkhorn}.
  To include a measure of ideal performance, we plot the {\margwass} metric between two independent sets of samples from the target with a black-dotted line (as in \Cref{fig:hmc_vs_flow_vi}).
  When an inference method's value approaches this line, its samples are as good as samples from the target.
  We use the same set of one million reference samples for all evaluations of a synthetic targets (\Cref{fig:hmc_vs_flow_vi,fig: batchsize and hmc}). In \Cref{sec:batch_size}, we also evaluate on some real targets and generate reference samples through long HMC runs (see \Cref{app: sec: details of batch_size} for details).
  
  \subsection{Experimental Details}
  \label{sec:exp details}
  To simulate different scales, we use three dimensions: two, ten, and one hundred (apart from Funana, where at least three dimensions are needed). 
  We implement flow-VI in JAX \citep{jax2018github} and use implementations in TensforFlow probability \citep{lao2020tfp} and NumPyro \citep{phan2019composable} for HMC methods. All methods are run on Nvidia A100 GPUs.  For full details,  see \Cref{app: sec: details of capacity,app: sec: details of divergences,app: sec: details of estimators,app: sec: details of optimization,app: sec: details of batch_size}.
  
  \section{DO REAL-NVP FLOWS HAVE ENOUGH CAPACITY FOR CHALLENGING TARGETS?}
  \label{sec:capacity}
  
Real-NVP flows are a popular choice among inference researchers \citep{webb2019improving,aagrawal2020,thin2020metflow,dhaka2021challenges,samsonov2022local,glockler2022variational,siahkoohi2023reliable,xu2023mixflows,andrade2024stable}.
However, some works also report unstable performance and poor results \citep{dhaka2021challenges,jaini2020tails,andrade2024stable}.
It is unclear whether such results arise due to a lack of representational ability of Real-NVP flows or a failure of optimization.
To resolve this, we study the impact of capacity while neutralizing other factors, asking: \emph{Do real-NVP flows have enough capacity to represent challenging targets accurately?} 
\begin{figure*}[ht!]
  \centering
  \includegraphics[trim = 0 0 0 0, clip, width = 0.85\textwidth]{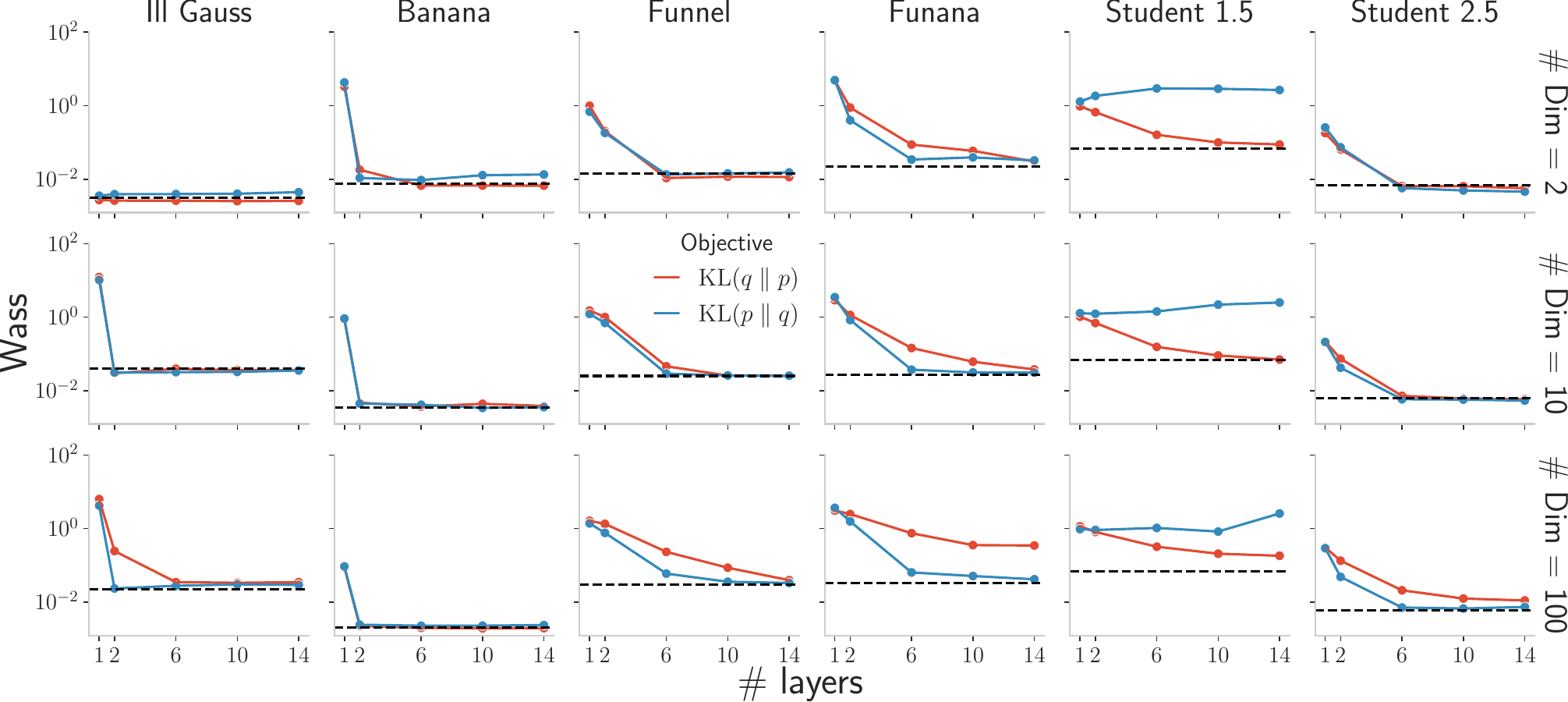}
  \caption{\small{\label{fig: objective wass} \textbf{Rows:} Model dimensions. {\Margwass} metric (\cref{eq: wass}) against the number of layers (with $32$ hidden units) for different objectives. Performance of $\KL{q}{p}$ (red) optimization improves as the number of layers increase, often reaching that of exact inference. A gap remains for Funana in one hundred dimensions, indicating this newly proposed density presents a significant challenge. (In the first row, Funana uses 3 dimensions.)}}
  \vspace{0pt}
\end{figure*}

% \jd{Need a much better argument for optimizing this divergence than "theoretically sound". I think our reasons are (a) that it makes optimization easier when we use exact samples and (b) we intuitively suspect it will be better? (Have other people suggested this)?}

% \jd{Beyond that, really need to organize thoughts in this paragraph more clearly. First explain what you want to do, then why, then how you did it.}

To answer the above question, we directly optimize $\KL{p}{q}$.
We do this by using a massive batch of samples from the target, significantly simplifying the optimization.
(The practical algorithms for $\KL{p}{q}$ optimization suffer from high-variance self-normalized importance sampling-based gradients \citep{geffner2020difficulty,geffner2021empirical}, but these issues get side-stepped when working with exact samples.)
Moreover, optimizing $\KL{p}{q}$ is generally preferable as the resulting approximations are expected to provide better coverage of the target compared to approximations obtained from optimizing the standard VI objective $\KL{q}{p}$ \citep{minka2005divergence,naesseth2020markovian,ou2020joint,kim2022markov}. 

We neutralize the impact of other optimization factors by sweeping over step-sizes and schedules, and optimizing for a large number of iterations (see \Cref{app: sec: details of capacity}).

% We optimize $\KL{p}{q}$, a theoretically sound objective \citep{minka2005divergence,naesseth2020markovian,ou2020joint,kim2022markov} conceptually equivalent to maximizing likelihood with infinite samples from the target.
% Optimizing $\KL{p}{q}$ is generally challenging due to unavailable exact samples and high-variance self-normalized importance sampling-based gradients \citep{geffner2020difficulty,geffner2021empirical}. 
% We sidestep these issues by directly optimizing $\KL{p}{q}$ using a massive batch of exact samples from the target.
% To remove the impact of other factors, we sweep over step-sizes and schedules, and optimize for a large number of iterations (see \Cref{app: sec: details of capacity} for details).

\Cref{fig: capacity} shows the {\margwass} metric as we vary the capacity by increasing the number of coupling layers or hidden units in the neural networks (see \Cref{app:sec:details of real-NVP} for details). The key insight is that with the appropriate choice of these architectural parameters, real-NVP flows have sufficient capacity to represent challenging targets. The one exception is the Student-t with $\nu = 1.5$, where $\KL{p}{q}$ optimization struggles due to heavy tails \citep{jaini2020tails} ($\KL{q}{p}$ performs well, see \Cref{fig: objective wass}). Key findings include the following.

\begin{enumerate}[leftmargin=15pt,itemsep=2pt,parsep=0pt,topsep=0pt,partopsep=0pt]
  \item \textbf{Layers:} Increasing the number of layers is very effective. With ten or more layers, performance often matches exact inference (curves get closer to the black dotted line as layers increase).
  \item \textbf{Hidden units:} Increasing the number of hidden units benefits higher dimensional problems (difference between the red and the blue curves increases).
\end{enumerate}

While one expects that increasing capacity improves representational power, these results suggest that existing flow models like real-NVP are capable of representing difficult posterior geometries. So, with appropriate optimization choices, one can hope for impressive results from ``relatively simple'' flows.

\section{DOES FLOW-VI NEED MODE-SPANNING OBJECTIVES?}
\label{sec:divergences}
The previous section demonstrates that real-NVP flows have enough capacity to represent challenging targets. 
However, \Cref{sec:capacity} relies on exact samples for optimization,
making it crucial to reconsider our  objective.
% Given that practical $\KL{p}{q}$ optimization methods suffer from high-gradient variance \citep{finke2019importance,geffner2020difficulty,geffner2021empirical}, 

Selecting the right optimization objective is an active research area. 
Some advocate ``mode-spanning'' objectives, like $\KL{p}{q}$ from \Cref{sec:capacity}, due to their better coverage \citep{li2016renyi,wang2018variational,dieng2017variational,hernandez2016black,margossian2024ordering,cai2024batch,zenn2024differentiable}.  
While these are promising, their practical algorithms can suffer from large gradient variance \citep{rainforth_tighter_2018,finke2019importance,geffner2021empirical,geffner2020difficulty}. 
The primary alternative is to use the ``easier'' standard VI objective. 
However, other works show that this also can struggle when used with flows \citep{behrmann2021understanding,dhaka2021challenges,andrade2024stable}. 
Overall, the choice remains unclear, raising questions like: \emph{Do we need complicated mode-spanning objectives? Does the standard VI objective suffice when the capacity is high?}
% The reasons for these results remain unclear due to confounding effects of different factors.

To answer the above questions, we focus on optimizing the ``easy'' $\KL{q}{p}$ while neutralizing the impact of other factors to see if it is sufficient for the excellent performance we saw in \Cref{sec:capacity} with $\KL{p}{q}$ optimization.
To do so, we use a massive batchsize (to reduce variance), do an exhaustive search over gradient estimators, step-sizes, and step-schedules, and run many iterations (see \Cref{app: sec: details of divergences} for details).

\begin{figure*}[ht!]
  \centering
  \includegraphics[width = 0.85\textwidth]{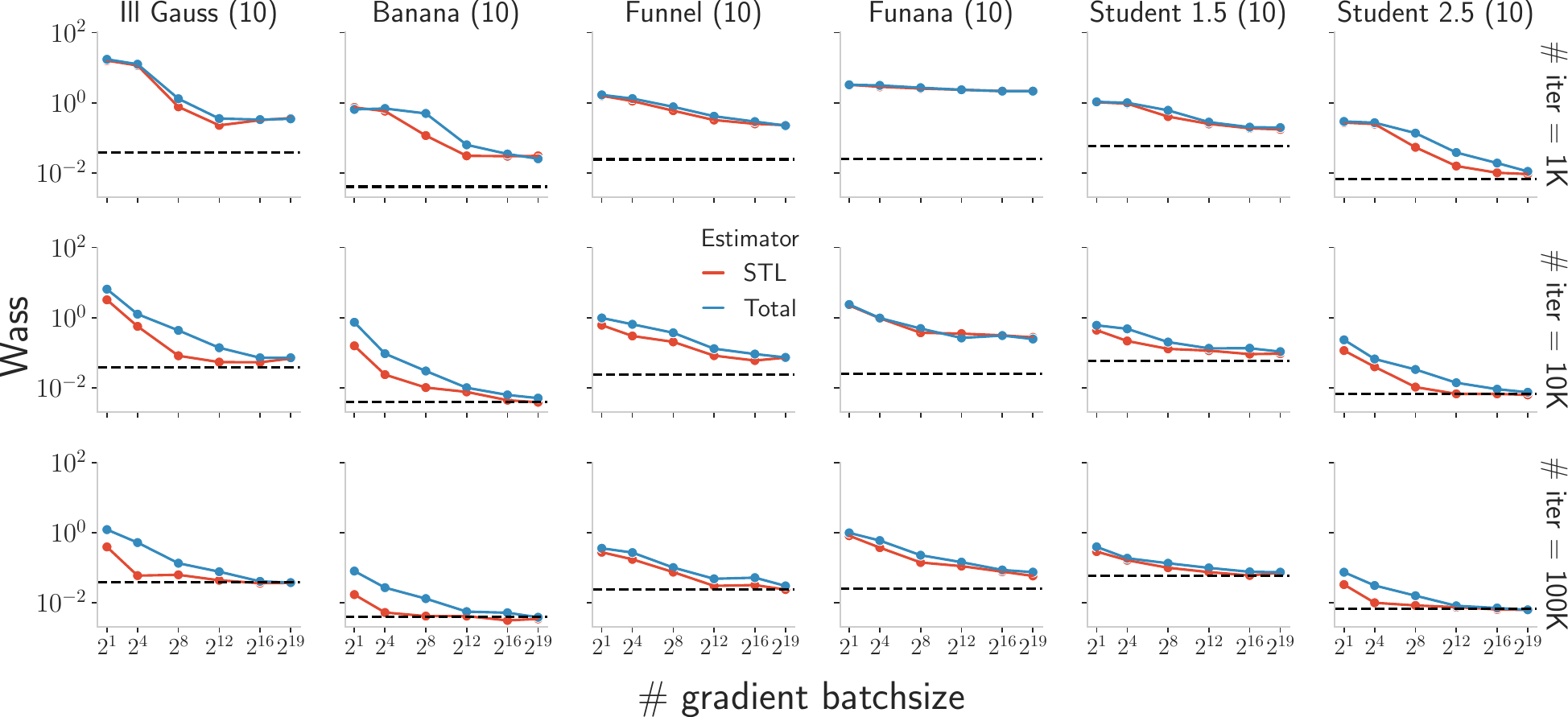}
\caption{\small{\label{fig: estimators} \textbf{Rows:} Number of iterations. {\Margwass} metric against the number of samples used for gradient evaluation for different targets in ten dimensions. STL (red) consistently outperforms total gradient (blue) at smaller batchsizes. However, as the batchsize increases, the difference vanishes. The performance for a given number of iterations (fixed for a row) improves significantly as the batchsize increases, indicating the impact of reduced gradient variance.}}
\vspace{0pt}
\end{figure*}

\Cref{fig: objective wass} plots {\margwass} metric as a function of the number of layers after optimizing $\KL{q}{p}$ and $\KL{p}{q}$. The key insights are:

\begin{enumerate}
[leftmargin=15pt,itemsep=2pt,parsep=0pt,topsep=0pt,partopsep=0pt]
  \item \textbf{Objective:} Optimizing the difficult "mode-spanning" $\KL{p}{q}$ (blue) indeed gives better results than the easier "mode-seeking" $\KL{q}{p}$ (red), except in the case of the Student-t with $\nu=1.5$, which is likely due to heavy tails \citep{jaini2020tails}.
  \item \textbf{Capacity:} As the capacity of the flow increases, the performance gap between the two objectives diminishes, with both achieving results close to exact inference. (A noticeable gap remains only for Funana in one hundred dimensions, suggesting this problem has a very challenging geometry).
\end{enumerate} 

These suggest an important difference for flow VI compared to VI with simple families like Gaussians. Rather than using a complex divergence to compensate for a mismatch between the target and the variational family, one can simply increase the flow capacity and optimize the standard objective. With enough capacity, the choice of divergence is less important.

\section{ARE REDUCED VARIANCE ESTIMATORS HELPFUL?}
\label{sec:estimators}

The previous sections showed that optimizing standard $\KL{q}{p}$ objective can approximate challenging targets if the flow has enough capacity. However, the strategy in \Cref{sec:divergences} relied on impractical exhaustive hyperparameter sweeps.  For effective practical use, we need a more efficient procedure. For this, we first ask: \emph{how to estimate the gradient for reliable optimization?}

One major challenge for optimizing ELBO (\cref{eq:ELBO}) is the reliance on stochastic gradient estimates. Several works have explored ways to construct lower-variance gradient estimators \citep{ranganath14,richter2020vargrad,geffner2020rule,miller2017reducing,roeder2017sticking,tucker2018doubly,bauer2021generalized,fujisawa2021multilevel,yi2023bridging,burroni2023u}.
A popular choice is the sticking-the-landing (STL) \citep{roeder2017sticking,tucker2018doubly} estimator, which often performs well \citep{aagrawal2020,dhaka2020robust,vaitl2022gradients,vaitl2024fast,andrade2024stable}. 
However, STL requires inverting the flow transformation $T$ \citep{aagrawal2020,vaitl2022gradients,vaitl2024fast}, adding computational cost and potential numerical issues \citep{behrmann2021understanding}. 
STL is also impractical for flows where inversion is expensive, like autoregressive flows \citep{kingma2016improved,papamakarios2017masked}.

In principle, modern GPUs allow a brute-force solution to the gradient variance problem\textemdash simply draw a huge number of estimates in parallel and average. This raises some natural questions: 
% \emph{When using high-capacity flows, how does the STL estimator interact with the batchsize?}
\emph{How does the choice of estimator interact with the batchsize when using high-capacity flows?}

% \jd{Do we really sweep over total number of iterations? Seems kinda odd. Is largest number not always selected?}
% \abhi{Where do we say we are sweeping over total number of iterations? }

To answer this question, we optimize a high-capacity flow with both STL and the standard total gradient estimator using different batchsizes. We run independent optimization for different numbers of iterations, and sweep over step-sizes and step-schedules to reduce the impact of optimization factors (see \Cref{app: sec: details of estimators}).

\Cref{fig: estimators} plots the {\margwass} metric against the gradient batchsize for targets in ten dimensions. The key insights are:
\begin{enumerate}[leftmargin=15pt,itemsep=2pt,parsep=0pt,topsep=0pt,partopsep=0pt]
  \item \textbf{Estimator:} STL (red) consistently outperforms the total gradient (blue) for small batchsizes, but the difference reduces as the batchsize increases, and both reach the accuracy of exact samples.
  \item \textbf{Batchsize:} For any fixed number of iterations (within a row), performance dramatically improves with larger batchsizes, highlighting the importance of reduced gradient variance.
\end{enumerate}

These findings highlight that
reducing gradient variance is crucial for strong empirical performance,
even for high-capacity flows that can represent challenging targets.
Large batchsizes significantly improve performance and should, almost certainly, be used when hardware is available to support computing them in parallel. STL also reduces variance and improves convergence and should usually also be used (when efficient, see \Cref{app:sec:details of stl for flows}).
\begin{figure*}[ht!]
  \centering
  \includegraphics[width = 0.85\textwidth]{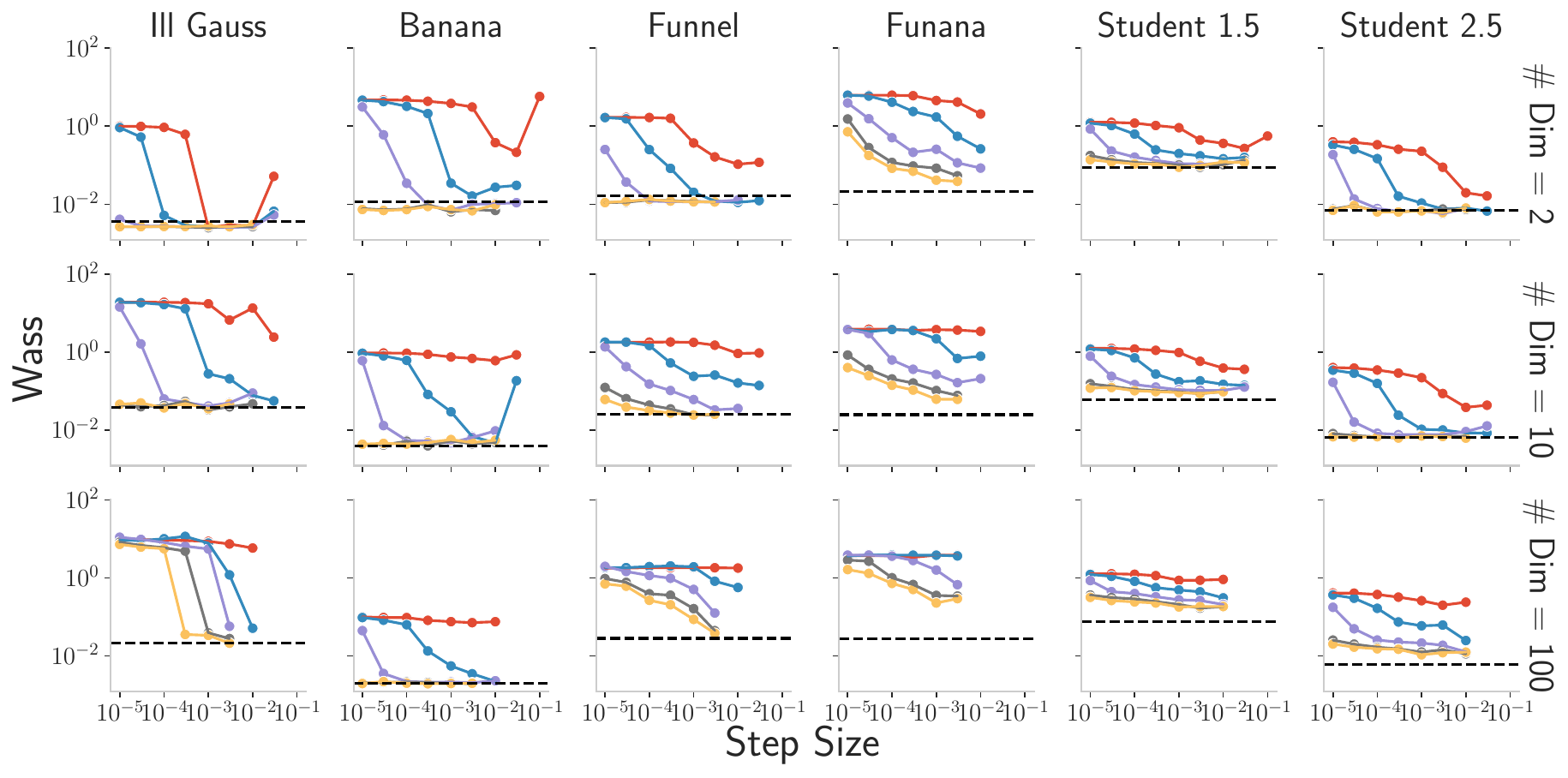}
  \includegraphics[width = 0.6\textwidth, trim = 10cm 0 15cm 15.2cm, clip]{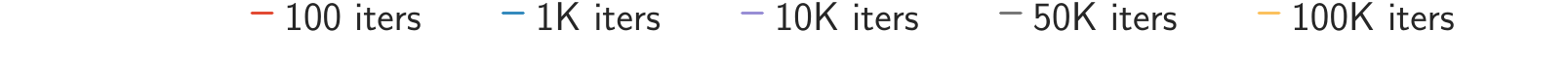}
  \caption{
    \small{\label{fig: optimization} \textbf{Rows:} Model dimensions. {\Margwass} metric versus step-sizes for different iteration counts. Each point represents an independent optimization run; missing points indicate divergence. Notably, certain step-sizes, like $10^{-2}$, achieve strong performance initially but diverge after several thousand iterations (see Funnel and Funana). 
    In general, step-sizes within the range $10^{-4}$ to $10^{-3}$ perform reliably across the targets when optimized for 10K iterations or more. 
    % \jd{whatever happened with the legend here is... unfortunate}}
    }}
  \vspace{0pt}
\end{figure*}

\section{CAN OPTIMIZATION BE RELIABLY AUTOMATED?}
\label{sec:optimization}

While the previous section focused on reliably estimating ELBO gradients, the remaining hyperparameter sweeps are still impractical. This leads to the next important question: \emph{how can we reliably optimize?}

Automating VI optimization is an open problem \citep{kucukelbir2017automatic,ambrogioni2021automatic_a,ambrogioni2021automatic_b,rouillard2023pavi,welandawe2024}. Decisions such as the choice of optimizer, step-size, step-schedule, and the number of iterations are not straightforward \citep{kucukelbir2017automatic,aagrawal2020}, and the literature offers little concrete guidelines tailored to flow VI.

Several works using flows report success with the Adam optimizer \citep{kingma2014adam}. Even when using Adam, selecting the optimal step-size schedule remains a challenge. This raises several questions: 
\emph{Is there a step-size range that works across the targets for high-capacity flows? Should we scale the step-size with the dimensionality? Can the best step-size be predicted based on early performance within a few hundred iterations?}

To transparently answer these questions, we optimize high-capacity flows independently with ten step-sizes for different numbers of iterations. We use a large batchsize and the STL estimator (see \Cref{app: sec: details of optimization}).

\Cref{fig: optimization} shows the {\margwass} metric against step-sizes. Some critical insights emerge:
\begin{enumerate}[leftmargin=15pt,itemsep=2pt,parsep=0pt,topsep=0pt,partopsep=0pt]
  \item \textbf{Diverging step-sizes:} Some step-sizes perform well for thousands of iterations and diverge when run for longer. For example, a step-size of \(10^{-2}\) works well for 1K iterations but diverges at 10K iterations in Funnel with one hundred dimensions.
  \item \textbf{Stable step-range:} Range of $10^{-4}$ to $10^{-3}$ consistently performs well across targets (Funnel in a hundred dimensions is a notable exception where a slightly larger step-size yields better results.)
\end{enumerate}

These results suggest that predicting optimal step-sizes from early performance (a few hundred iterations) is challenging. When using high-capacity flows with low gradient variance, step-sizes within a narrow range from $10^{-4}$ to $10^{-3}$ perform optimally across different targets and dimensions when optimized for at least 10K iterations (see purple, gray, and yellow curves).

\section{HOW DOES FLOW VI COMPARE TO HMC METHODS?}
\label{sec:batch_size}
\begin{figure*}[ht!]
  \centering
  \includegraphics[width = 0.85\textwidth, trim={0, 2.1cm, 0, 0}, clip]{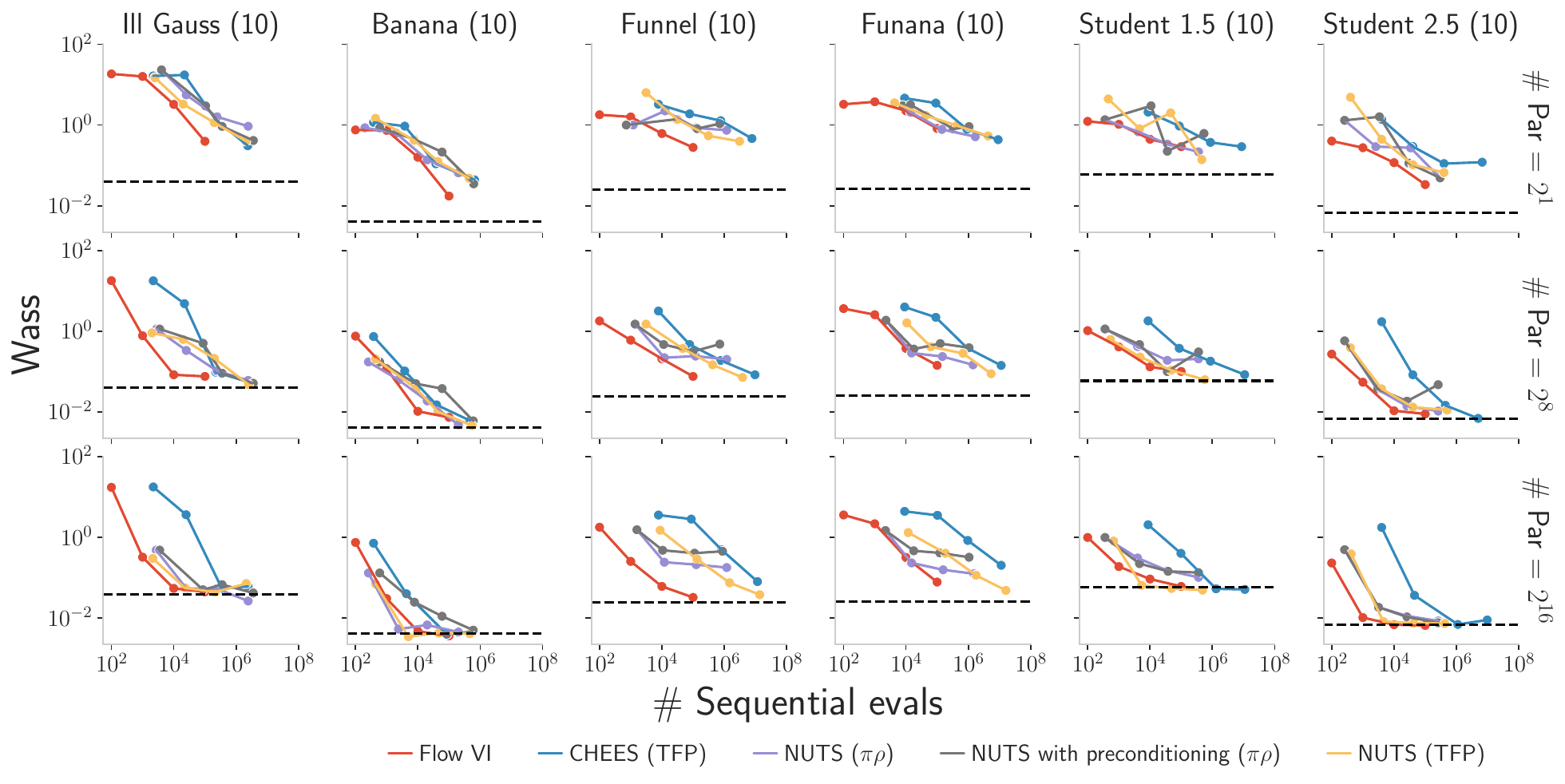}\vspace{4pt}
  \includegraphics[width = 0.85\textwidth, trim={0, 1.65cm, 0, 0}, clip]{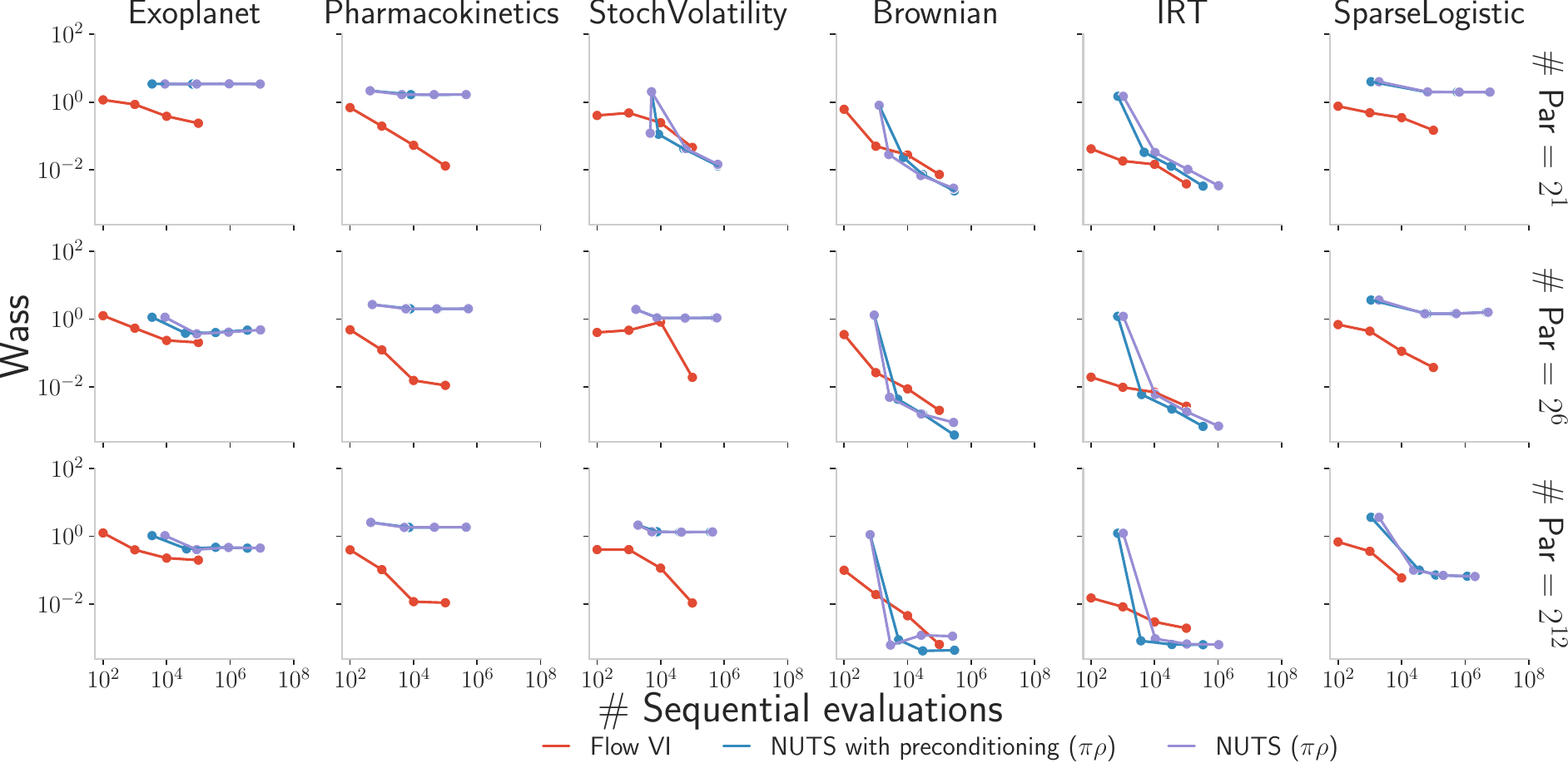}
  \includegraphics[width = 0.65\textwidth, trim={4cm, 1cm, 4cm, 14.5cm}, clip]{./figures/section_7/new_final_scaled_wass_values_all_methods_all_models_ndim_10_batchsizes____2__1___,___2__8___,___2__16_____all_hmc.pdf}\\
  \includegraphics[width = 0.65\textwidth, trim={5.5cm, 0, 1.5cm, 15.75cm}, clip]{./figures/section_7/new_final_scaled_wass_values_all_methods_all_models_ndim_10_batchsizes____2__1___,___2__8___,___2__16_____all_hmc.pdf}\\
  \caption{\small{\label{fig: batchsize and hmc} 
  % For flow VI, the parallel evaluations corresponds to gradient batchsize and the sequential evaluation correspond to number of optimization iterations. For HMC, parallel evaluations represents the number of chains and sequential evaluations represent the number of leapfrog steps.  (We use implementations from NumPyro ($\pi\rho$) and TensforFlow Probability (TFP) for HMC methods, see \Cref{app: sec: details of batch_size} for details.) 
  % All the methods improve as the amount of parallel compute increases. 
  \textbf{Rows:} Parallel evaluations. {\Margwass} metric against number of sequential evaluations for (a) synthetic targets in ten dimensions and (b) real-world problems, with parallel evaluations increasing across the rows. 
  With sufficient parallel compute, flow VI is almost as accurate as NUTS (or exact inference); however, flow VI takes much fewer number of sequential evaluations.
  See \Cref{fig:hmc_vs_flow_vi} and \Cref{sec:batch_size} for details on how the evaluations are computed.
  }}
  \vspace{0pt}
\end{figure*}

At this point, previous sections suggest a simple recipe: Use a high capacity flow, traditional VI objective (for easier optimization), the STL estimator (when possible, for lower gradient variance), a large gradient batchsize (to further lower the variance), and a fixed step-size in a small range  (see \Cref{app: sec: recipe} for detailed recommendations).
Naturally, we ask: \emph{how well does this actually work?} In this section, we compare our recipe to the state-of-the-art turnkey HMC methods.

% \jd{We blur the line between HMC and NUTS in this paper. Usually that's OK-ish, but it's a problem in the paragraph below as the control flow issues only really arise with NUTS.}

% \jd{Shold mention that CHEES is designed to try to be better with parallel compute}

Using modern GPUs with flow VI is straightforward—simply average more stochastic gradient estimates in each iteration. 
However, leveraging them for HMC methods is challenging \citep{margossian2023many,hoffman2021adaptive,hoffman2022tuning}. 
For NUTS, a leading turnkey HMC method, complications arise due to nuanced chain-dependent control flows \citep{lao2019unrolled,phan2019iterative,radul2020automatically,hoffman2021adaptive}.
For some recent methods, like CHEES \citep{hoffman2021adaptive} or MEADS \citep{hoffman2022tuning}, the need for efficient cross-chain communication complicates implementations. 
Currently, running multiple parallel chains is the simplest and most effective way to use GPUs with HMC methods \citep{margossian2021nested,hoffman2021adaptive}. 
% For our comparisons, we consider NUTS and CHEES.

%  and use implementations from TensorFlow Probability \citep{lao2020tfp} and NumPyro \citep{phan2019composable}.

With real problems, the computational bottleneck is often the model evaluation. Therefore, we count the model evaluations along two axes: parallel (a proxy for accelerator size) and sequential (a proxy for runtime). For HMC, parallel evaluations correspond to the number of chains, and sequential evaluations are the number of leapfrog steps. For VI, parallel evaluations correspond to gradient batchsizes, and sequential evaluations are the number of optimization iterations. Due to the limitations of the current frameworks, we extrapolate warmup-phase leapfrog-steps from the post-warmup phase. (See \Cref{app: sec: details of batch_size}.)

\Cref{fig: batchsize and hmc} plots the {\margwass} metric against the number of sequential evaluations for different target densities. 
In addition to synthetic targets, we also include comparisons on six real-world problems (see \Cref{app: sec: details on other models} for model details).
Key insights are:
\begin{enumerate}[leftmargin=15pt,itemsep=2pt,parsep=0pt,topsep=0pt,partopsep=0pt]
\item \textbf{Parallel evals:} Performance for all methods improves significantly as parallel compute increases (moving top to bottom across the rows).% (as we go top to bottom row, all methods achieve lower {\margwass}).
\item \textbf{Sequential evals:} Flow VI matches or surpasses HMC methods for smaller batchsizes, and requires far fewer number of sequential evaluations when the batchsizes are larger.
\end{enumerate}  

These results suggest that flow VI can more effectively utilize large batchsizes, avoiding the coordination overhead inherent in HMCs chain-based approaches. More advances in HMC for modern accelerators could temper this conclusion; however, for now, flow VI using our proposed recipe serves as a strong alternative. 
% \begin{figure*}
%   \centering
%   \includegraphics[width = 0.8\textwidth]{./figures/section_7/real_models_without_correction_[2, 64, 4096]_all_models.pdf}
%   \caption{\small{
%       \label{fig: real models} \textbf{Rows:} Parallel evaluations. {\Margwass} metric against number of sequential evaluations for different models. The evaluations are for Flow VI and HMC methods are computed as described in \Cref{fig: batchsize and hmc}. With sufficient parallel compute (last row), flow VI is almost as accurate as NUTS (and sometimes better); however, takes much fewer number of sequential evaluations.
%   }}
% \end{figure*}

\section{CONCLUSION}
\label{sec:discussion}

This paper presents a step-by-step analysis to disentangle the impact of key factors of flow VI. 
Our analysis finds that high-capacity flows and large gradient batchsizes are essential for achieving strong performance. 
We provide recommendations for successful choices of objectives, gradient estimators, and optimization strategies. 
Additionally, we show flow VI can match or surpass leading turnkey HMC methods on challenging targets with much fewer sequential steps. 

  \bibliography{paper}
  \bibliographystyle{plainnat}

\section*{Checklist}

\begin{enumerate}

  \item For all models and algorithms presented, check if you include:
  \begin{enumerate}
    \item A clear description of the mathematical setting, assumptions, algorithm, and/or model. [Not Applicable]
    \item An analysis of the properties and complexity (time, space, sample size) of any algorithm. [Not Applicable]
    \item (Optional) Anonymized source code, with specification of all dependencies, including external libraries. [No]
  \end{enumerate}

  \item For any theoretical claim, check if you include:
  \begin{enumerate}
    \item Statements of the full set of assumptions of all theoretical results. [Not Applicable]
    \item Complete proofs of all theoretical results. [Not Applicable]
    \item Clear explanations of any assumptions. [Not Applicable]     
  \end{enumerate}

  \item For all figures and tables that present empirical results, check if you include:
  \begin{enumerate}
    \item The code, data, and instructions needed to reproduce the main experimental results (either in the supplemental material or as a URL). [Yes`']
    \item All the training details (e.g., data splits, hyperparameters, how they were chosen). [Yes]
          \item A clear definition of the specific measure or statistics and error bars (e.g., with respect to the random seed after running experiments multiple times). [Yes] 
          \item A description of the computing infrastructure used. (e.g., type of GPUs, internal cluster, or cloud provider). [Yes]
  \end{enumerate}
 
  \item If you are using existing assets (e.g., code, data, models) or curating/releasing new assets, check if you include:
  \begin{enumerate}
    \item Citations of the creator If your work uses existing assets. [Yes]
    \item The license information of the assets, if applicable. [Not Applicable]
    \item New assets either in the supplemental material or as a URL, if applicable. [Not Applicable]
    \item Information about consent from data providers/curators. [Not Applicable]
    \item Discussion of sensible content if applicable, e.g., personally identifiable information or offensive content. [Not Applicable]
  \end{enumerate}
 
  \item If you used crowdsourcing or conducted research with human subjects, check if you include:
  \begin{enumerate}
    \item The full text of instructions given to participants and screenshots. [Not Applicable]
    \item Descriptions of potential participant risks, with links to Institutional Review Board (IRB) approvals if applicable. [Not Applicable]
    \item The estimated hourly wage paid to participants and the total amount spent on participant compensation. [Not Applicable]
  \end{enumerate}
 
  \end{enumerate}
 \endgroup
 \onecolumn
\appendix

\renewcommand{\contentsname}{Appendices: Table of Contents} % Change title
\tableofcontents  

\section{RELATED WORKS}
\label{app:sec:related works}
% \textbf{Related Works.}
Several works explore normalizing flows for black-box variational inference (BBVI) \citep{webb2019improving,baudart2021automatic,aagrawal2020,dhaka2021challenges,ambrogioni2021automatic_b,andrade2024stable}.
Often, these studies make axiomatic assumptions about flow abilities of flows and are more focused on the applications. 
We focus on disentangling the key factors involved in flow VI using a step-by-step approach and our learnings should benefit any such future applications. 

Numerous studies aim to automate aspects of BBVI \citep{kucukelbir2017automatic,aagrawal2020,dhaka2020robust,ambrogioni2021automatic_a,ambrogioni2021automatic_b,welandawe2024}, aiming to provide turnkey solutions for probabilistic models without requiring manual intervention. 
Unlike these approaches, our work focuses on understanding the impact of different factors, offering detailed guidelines rather than an automatic solution. We believe our analysis lays the necessary groundwork for future flow-based automatic inference tools. 

\citet{aagrawal2020} proposed an approach that combines normalizing flows, STL \citep{roeder2017sticking}, a step-size search scheme, and a post-hoc importance sampling step to enhance out-of-the-box BBVI performance. 
They do not dissect the factors affecting flow VI, use a relatively small batchsizes, use CPUs, and skip any insights into their failure cases. In contrast, our work specifically delves into the impact of different factors, leverages modern GPUs to employ massive batchsizes, and also demonstrates how appropriately optimized flow VI can match or surpass HMC methods.

\citet{andrade2024stable} focus on stably optimizing real-NVP flows on high-dimensional problems and consider the impact of architectural choices on optimization stability. 
Our study independently uses an architecture that aligns with their optimal choices, avoiding instabilities. While it is possible that some of their tricks improve our performance, our aim here is to develop a holistic understanding of impact of different factors and not just restrict to architectural choices. 
 
\citet{dhaka2021challenges} use an importance sampling based diagnostics to assess performance and recommend using normalizing flows but report poor performance due to optimization challenges (see Figures C.2 and C.3 in their appendix). Based on the insights from our study, we conjecture that the relatively lower-capacity flows (causing insufficient representational ability) and smaller batchsizes (leading to higher-gradient variance) might explain some issues they experience. 

\citet{jaini2020tails} uncover that targets with heavier tails require approximations that have base distributions with identical tail properties. Our findings in \Cref{fig: capacity} corroborate this as optimizing $\KL{p}{q}$ results in poor performance on heavy tailed targets. Importantly, in \Cref{sec:divergences}, a high-capacity flow with standard VI objective significantly improves the performance.

\citet{blessing2024beyond} highlighted the need for standardized evaluation in inference research and introduced a benchmark that includes both synthetic and real-world problems with special focus on multi-modal targets.
We also use synthetic densities to create controlled environments and employ integral metrics like the Wasserstein distance. 
However, we additionally provide a useful measure of ideal performance to better contextualize the numbers (black dotted line in figures, see \Cref{sec:eval}). 
We also include a collection of six real-world problems to showcase the practical performance of flow VI.

Recent literature also explores the use of flows as proposals for MCMC methods \citep{parno2018transport,hoffman2019neutra,wu2020stochastic,naesseth2020markovian,arbel2021annealed,matthews2022continual,gabrie2022adaptive,hagemann2022stochastic,brofos2022adaptation,kim2022markov,samsonov2022local,cabezas2023transport} with some using $\KL{q}{p}$ optimization to initialize the proposal distribution.  
These approaches are orthogonal to simply using flow VI, and can potentially benefit from insights discovered in this work (by learning better initial proposals).

\section{SUGGESTED RECIPE}
\label{app: sec: recipe}

Flow VI requires several careful choices for great performance.
However, some of these depend on the available computational resources.
To facilitate easier adoption, we suggest a \emph{recipe} and encourage practitioners to adapt this to best fit their constraints. 
% Specifically, we recommend the following.

\textbf{Absolute essentials.} Usually this will be easier to accommodate with powerful GPUs. 
\begin{itemize}[leftmargin=15pt]
  \item {Capacity:} Use high capacity flows to reduce representational constraints. We use real-NVP flows with at least ten coupling layers and 32 hidden units when not varying the capacity explicitly (see \Cref{fig: objective wass,fig: capacity}, and \Cref{app:sec:details of real-NVP} for architectural details).
  \item {Batchsize:} Use a large number of gradient estimates (batchsize) to lower the gradient variance and simplify optimization. We use $512 \times 2^{10}$ samples for experiments that do not explicitly vary the batchsize. However, performance is good with even smaller choices (see \Cref{fig: estimators}). 
\end{itemize}
\textbf{Works really well.} With high capacity and large batchsize, the following recommendations work extremely well.
% When working with above essential choices of high-capacity and large batchsize, these choices work extremely well.
\begin{itemize}[leftmargin=15pt]
  \item {Objective:} Optimize traditional VI objective. While, mode-spanning objectives help, the standard objective is easier to optimize and achieves comparable performance (see \Cref{fig: objective wass}).
  \item {Estimator:} Use sticking-the-landing (STL) gradient estimator to reduce gradient variance. For some flows, this can be tricky, see \Cref{app:sec:details of stl for flows} for a discussion.
  \item {Step-size:} Select a fixed step-size in $10^{-4}$ to $10^{-3}$ (\Cref{fig: optimization}). If resources allow, do a sweep within this range. 
  \item {Optimization:} Optimize for many iterations with an adaptive optimizer like Adam. At least ten thousand updates worked extremely well across the targets and dimensions (see \Cref{fig: optimization}).
\end{itemize}
\textbf{Additional ingredients.} Some additional recommendations based on initial experimentation. 
\begin{itemize}[leftmargin=15pt]
  \item {Base distribution:} Evaluate pre-optimization ELBO under different distributions and choose the one with the highest initial ELBO. For transparent experiments, we use the standard normal for the base distribution. However, for several targets, setting the Laplace's approximation as the base distribution can help warm start optimization \citep{domke2018importance,domke2019divide,aagrawal2020}. Setting a Student-t distribution as the base distribution has also been shown to achieve impressive results for heavy-tail targets \citep{jaini2020tails,liang2022fat,andrade2024stable}. We suggest doing a small check before optimization and picking the one with the higher value. 
  \item {Parameter initialization:} Initializing the neural network parameters to some small value is equivalent to initializing the flow-transformations to identity \citep{aagrawal2020}. We suggest using this for greater control over the starting distribution.
  \item {Non-linearity for scale function:} Chose hyperbolic tangent for the scale function in the affine transform (see \cref{eq:rnvp-appendix} and related discussion in \Cref{app:sec:details of real-NVP}). The choice of this non-linearity impacts the stability of the affine coupling flows \citep{behrmann2021understanding,andrade2024stable}. We experimented with several scale non-linearities and found that the hyperbolic tangent wrapped in an exponential (as in \cref{eq:rnvp-appendix}) provides an easy trade-off of stability and capacity that performs consistently across the targets. 
\end{itemize} 

\section{DETAILS OF $n$-WASSERSTEIN EVALUATION}
\label{app: sec: details of wasserstein evaluation.}
\begin{figure}[ht!]
  \centering
  \begin{subfigure}{\linewidth}
    \centering
    \includegraphics[width = 0.8\textwidth]{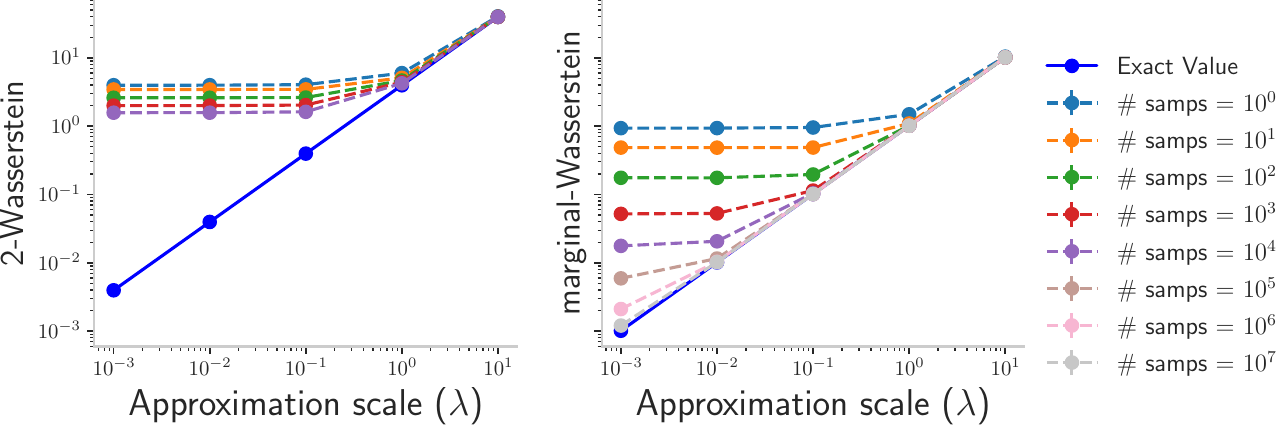}
    \captionsetup{justification=centering}
    \caption{\label{fig: wasserstein empirical a}
    % \small{Empirical and closed-form marginal-Wasserstein and $2-$Wasserstein values.}
    }
    \vspace{4pt}
  \end{subfigure}
  \hfill
  \begin{subfigure}{0.8\linewidth}
    \centering
    \includegraphics[width = 0.8\textwidth]{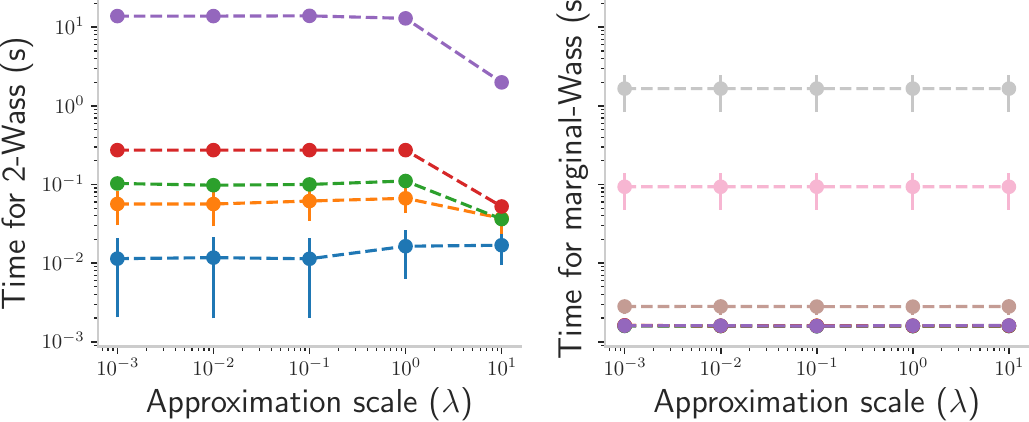}
    \captionsetup{justification=centering}
    \caption{\label{fig: wasserstein empirical b}
    % \small{Time taken to compute empirical marginal-Wasserstein and $2-$Wasserstein.}
    }
  \end{subfigure}
  \caption{\label{fig: wasserstein empirical demonstration} \small{\textbf{(a)} Empirical $2-$Wasserstein and {\margwass} metric against approximation scale $\lambda$ for the example described in \Cref{app: sec: example for wasserstein calculations}. As $\lambda$ increases, approximation becomes poor, increasing the true value of either of the Wasserstein metrics (solid blue line). For poor approximations (large $\lambda$), the numerical evaluation is accurate with smaller sample-size. However, when the approximation is closer to the target (small $\lambda$), a larger sample-size is required for accurate evaluation and fine comparisons. \textbf{(b)} Time taken for calculating $2-$Wasserstein and {\margwass} metric against the approximation scale $\lambda$. {\Margwass} scales log-linearly (\Cref{sec:wass}), allowing much larger sample-sizes while maintaining reasonable wall-clock times. (Calculating $2-$Wasserstein with $10^{4}$ samples take more time than $10^{7}$ samples for {\margwass} metric).}}
\end{figure}

Wasserstein distance between two distributions is generally not available in closed-form \citep{villani2009optimal}.
Numerical methods to evaluate it do not scale well with sample size \citep{cuturi2013sinkhorn}. 
When using a limited number of samples, these lead to noisy evaluations that make the $n-$Wasserstein distance unreliable for fine-grained comparisons. 
To demonstrate this, we present a simple representative example. 

\subsection{Example for Wasserstein Calculations} 
\label{app: sec: example for wasserstein calculations}
Consider a standard normal target distribution in ten dimensions. We will evaluate $2-$Wasserstein and the {\margwass} metric between this target and some approximations. Both of the Wasserstein metrics are available in closed form for Gaussian distributions \citep{djalilwass,chhachhi20231}. We will compare this exact value with numerically calculated values to understand the efficacy of these metrics for making fine-grained comparisons.

To create approximation that are increasingly different from the target (the standard normal), we sample the location from  $\mathcal{N}(0, \lambda^{2} \mathbf{I})$ where we vary the approximation scale $\lambda$ geometrically from $10^{-4}$ to $10^{1}$. When $\lambda$ is small (like $\lambda = 10^{-4}$), the approximation is close to the target, and when $\lambda$ is large (like $\lambda = 10$), the approximation is likely to be very different from the target. (The covariance matrix is set to be identity for all the approximations.)

\Cref{fig: wasserstein empirical demonstration} shows the performance of empirical calculation of {\margwass} metric and $2-$Wasserstein. 
When the approximation is poor ($\lambda$ is large), the numerical methods require a much smaller sample-size for accurate estimation (\Cref{fig: wasserstein empirical a}). 
However, when the approximation is close to the target ($\lambda$ is small), both of these metrics require a much larger sample-size for accurate estimation. Since, {\margwass} metric is more scalable, we are able to use a much larger sample-size and get extremely accurate measurements in relatively faster wall-clock time (see \Cref{fig: wasserstein empirical b}). We use \texttt{ott-jax} \citep{cuturi2022optimal} for calculating the $2-$Wasserstein distance and run the experiments on Nvidia A100.

Overall, \Cref{fig: wasserstein empirical demonstration} demonstrates that unless we use a large number of samples, approximations with varying degree of accuracy can evaluate to the same incorrect Wasserstein value. 
$2-$Wasserstein metric is not suitable for evaluating fine-grained comparisons as it can not scale to large sample sizes required for such accurate evaluations. On the other hand, {\margwass} metric is a lot more scalable and provides accurate measurements in reasonable time.

\section{LIMITATIONS}
\label{app:sec:limitations}
We use synthetic targets in our setup. Our use is intentional to mitigate any possible variability and to facilitate high-fidelity performance comparisons with HMC methods. 
Future work can exhaustively explore the impact on real-world problems. However, we expect our learnings to apply. In fact, we show that flow VI can match or surpass HMC methods on several real-world problems \Cref{fig: batchsize and hmc,app: fig: real models}.
We use modern GPUs like the Nvidia A100s which are not yet universally available. Despite computational constraints, we believe researchers and practitioners can still take guidelines we establish to better utilize the resources available to them.
We do not go beyond a hundred dimensions for the synthetic targets. While this is not small, our step-by-step analysis required a substantial computational effort, consuming more than 4000 GPU hours on NVIDIA A100s. We expect several of our learnings to apply when scaling to even higher dimensions, and leave the in-depth exploration for future work. 
Some expert practitioners may find some of our findings trivial. However, we believe the utility of these techniques is hypothetical unless somebody conducts a direct, thorough study like we do.

\section{DETAILS OF REAL-NVP ARCHITECTURE}
\label{app:sec:details of real-NVP}
We use a  real-NVP \citep{dinh2016density} flow with affine coupling layers. We define each coupling layer to be comprised of two transitions, where a single transition corresponds to affine  transformation of one part of the latent variables. For example, if the input variable for the $k^{th}$ layer is $z^{(k)}$, then first transition is defined as     
\begin{align}  
  z_{1:d} &= z^{(k)}_{1:d} \quad \text{and}\nonumber\\
  z_{d+1:D} &= z^{(k)}_{d+1:D} \odot \exp\big(s^{a}_{k}(z^{(k)}_{1:d})\big) + t^{a}_{k}(z^{(k)}_{1:d}),
  \label{eq:rnvp-appendix}
\end{align}  
where, for the function $s$ and $t$, super-script $a$ denotes first transition and sub-script $k$ denotes the layer $k$. For the next transition, the $z_{d+1:D}$ part is kept unchanged and $z_{1:d}$ is affine transformed similarly to obtain the layer output $z^{(k+1)}$ (this time using $s^{b}_{k} (z^{(k)}_{d+1:D}) $ and $t^{b}_{k} (z^{(k)}_{d+1:D}) $). 
This is also referred to as the alternating first half binary mask. 

Both, scale($s$) and translation($t$) functions of single transition are parameterized by the same fully connected neural network(FNN). More specifically, in the above example, a single FNN takes $z^{(k)}_{1:d}$ as input and outputs both $s^{a}_{k}(z^{(k)}_{1:d})$ and $t^{a}_{k}(z^{(k)}_{1:d})$. The skeleton of the FNN, in terms of the size of the layers, is as $[d, H,H, 2(D-d)]$ where, $H$ denotes the size of the two hidden layers.  

The hidden layers of FNN use a leaky rectified linear unit with slope = 0.01, while the output layer uses a hyperbolic tangent for $s$ and remains linear for $t$. 
We initialize neural network parameters with normal distribution $\mathcal{N}(0, 0.001^{2})$. This choice approximates standard normal initialization.

\section{DETAILS OF STL FOR FLOWS}
\label{app:sec:details of stl for flows}
% We use normalizing flows \citep{rezende2015variational,kobyzev2020normalizing,papamakarios2019normalizing} as the variational family for $q$. The main idea behind flows is to transform a base density $q_\epsilon(\epsilon)$ using a diffeomorphism $T$. Let $\epsilon \sim q_\epsilon(\epsilon)$. Then, the transformed variable $z = T(\epsilon)$ has the density
% $q(z) = q_\epsilon(\epsilon) |\det \nabla_\epsilon T(\epsilon)|^{-1}.$ Usually, the transformation $T$ is composed of a sequence of neural-network based transformations designed such that $T$ is invertible and the determinant of the Jacobian $|\det \nabla_\epsilon T(\epsilon)|$ can be efficiently calculated \citep{dinh2014nice,rezende2015variational,dinh2016density,kobyzev2020normalizing,papamakarios2019normalizing}.

To understand the issue with implementing sticking-the-landing (STL) \citep{roeder2017sticking,tucker2018doubly} estimator for flows, it is helpful to keep two fundamental steps in mind. 
First, generate a sample $z$ from $q$. 
Second, evaluate the density of a sample $z$ under $q$. 
Both of these steps are integral to estimate ELBO and estimating its gradients.
In most flows, these two steps can be carried out in a single forward pass by keeping track of the Jacobian calculation $|\det \nabla_\epsilon T(\epsilon)|$ (to evaluate the density) while transforming input samples $\epsilon$ (to generate the sample). 

However, a single-forward pass is insufficient for correct STL calculations and extra care is required.
To see why, note the main idea of path-gradients (like STL) is to restrict the dependence of learnable parameters to only the sampling step and to ignore any dependence during density evaluation.
To implement this, we need to manipulate the flow of gradients such that the updates only depend on the sampling step.
This requirement of treating the parameters differently during these two fundamental steps of sampling and evaluation creates the implementation challenge that forbids a single forward-pass evaluation.

\citet{aagrawal2020} implement STL for flows by implementing the two steps separately. This naive implementation increases memory and almost doubles the run-time. Recent work has improved the STL implementation for flows both in terms of memory and speed, but requires more complicated implementation procedures \citep{vaitl2022gradients,vaitl2024fast}.
All of these approaches still require both forward and inverse passes of flow to be efficient \citep{vaitl2024fast}.
Thus, despite these advances, autoregressive flows \citep{kingma2016improved,papamakarios2017masked} still can not be practically used with STL as they are only efficient in one direction.

\section{DETAILS FOR \Cref{sec:capacity} EXPERIMENTS}
\label{app: sec: details of capacity}
We directly optimize $\KL{p}{q}$ using samples from the target density. We use a gradient batchsize of $512\times2^{10}$ samples, that is, at each iteration, we draw a fresh batch of $512\times 2^{10}$ samples from the target density to approximate the $\KL{p}{q}$ gradients. 
We optimize using Adam \citep{kingma2014adam} and sweep over three step-sizes: $0.0001, 0.0003, $ and $ 0.001$, and two step-schedules: decayed and constant. 
We run the optimization for $100$K iterations. 

Decayed schedule scales down the step-size by a factor of ten, twice during the optimization: once, after half the number of iterations and again after three-fourth number of iterations. For example, if we start with a step-size of $0.01$, the step-size is kept constant for the first half and then updated at the half-way point to $0.001$. Then, we keep it constant for the next one-fourth of iterations at $0.001$ and update it at the three-fourth point to $0.0001$.

We vary the capacity by changing the number of hidden units in the two-hidden layer FNN and the number of coupling layers as described in \Cref{app:sec:details of real-NVP}. Each coupling layer corresponds to two affine transformations, resulting in a complete transformation of the input variables (\Cref{app:sec:details of real-NVP} for details).
For hidden units, we use $2$ and $32$, and for coupling layers, we use $1, 2, 6, 10,$ and$ 14$.

\section{DETAILS FOR \Cref{sec:divergences} EXPERIMENTS}
\label{app: sec: details of divergences}
For $\KL{q}{p}$ optimization, we use a gradient batchsize of $512\times2^{10}$ samples, that is, at each iteration, we draw a fresh batch of $512\times 2^{10}$ samples from $q$ to approximate the $\KL{q}{p}$ gradients. 
We use Adam for optimization and sweep over gradient estimators: STL and total gradient, three step-sizes: $0.0001, 0.0003,$ and $0.001$ and two step-schedules: decayed and constant. 
We optimize for $100$K iterations for all these choices. 

For capacity and $\KL{p}{q}$ results, please see the details in \Cref{app: sec: details of capacity}.
\section{DETAILS FOR \Cref{sec:estimators} EXPERIMENTS}
\label{app: sec: details of estimators}
\begin{figure}[ht!]
  \centering
  \begin{subfigure}{\linewidth}
    \centering
    \includegraphics[width = 0.8\textwidth]{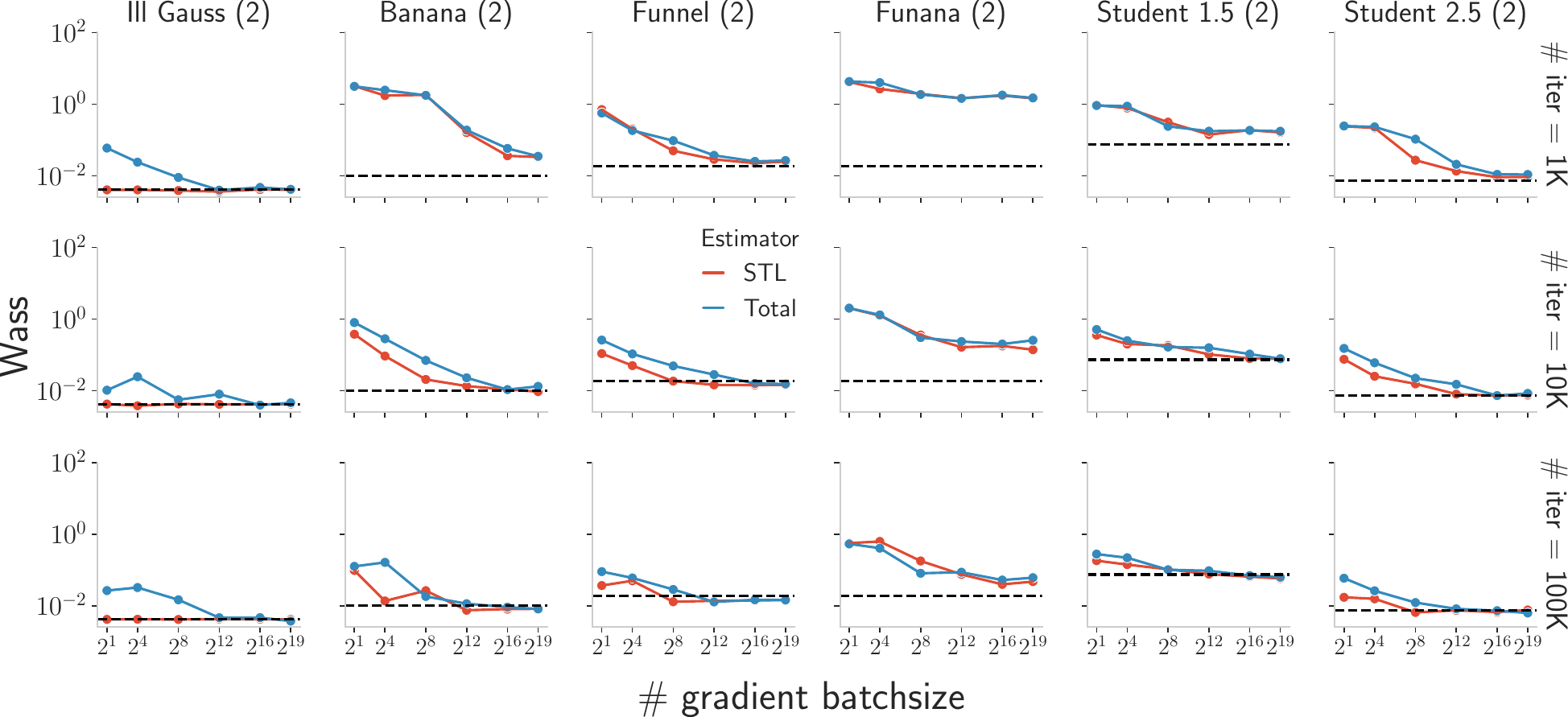}
    \captionsetup{justification=centering}
    % \caption{Results in 2 dimensions}
  \end{subfigure}
  \begin{subfigure}{\linewidth}
    \centering
    \includegraphics[width = 0.8\textwidth]{./figures/section_5/final_scaled_wass_values_all_methods_all_models_ndim_10_iters___1K_,__10K_,__100K___all.pdf}
    \captionsetup{justification=centering}
    % \caption{Results in 10 dimensions}
  \end{subfigure}
  \begin{subfigure}{\linewidth}
    \centering
    \includegraphics[width = 0.8\textwidth]{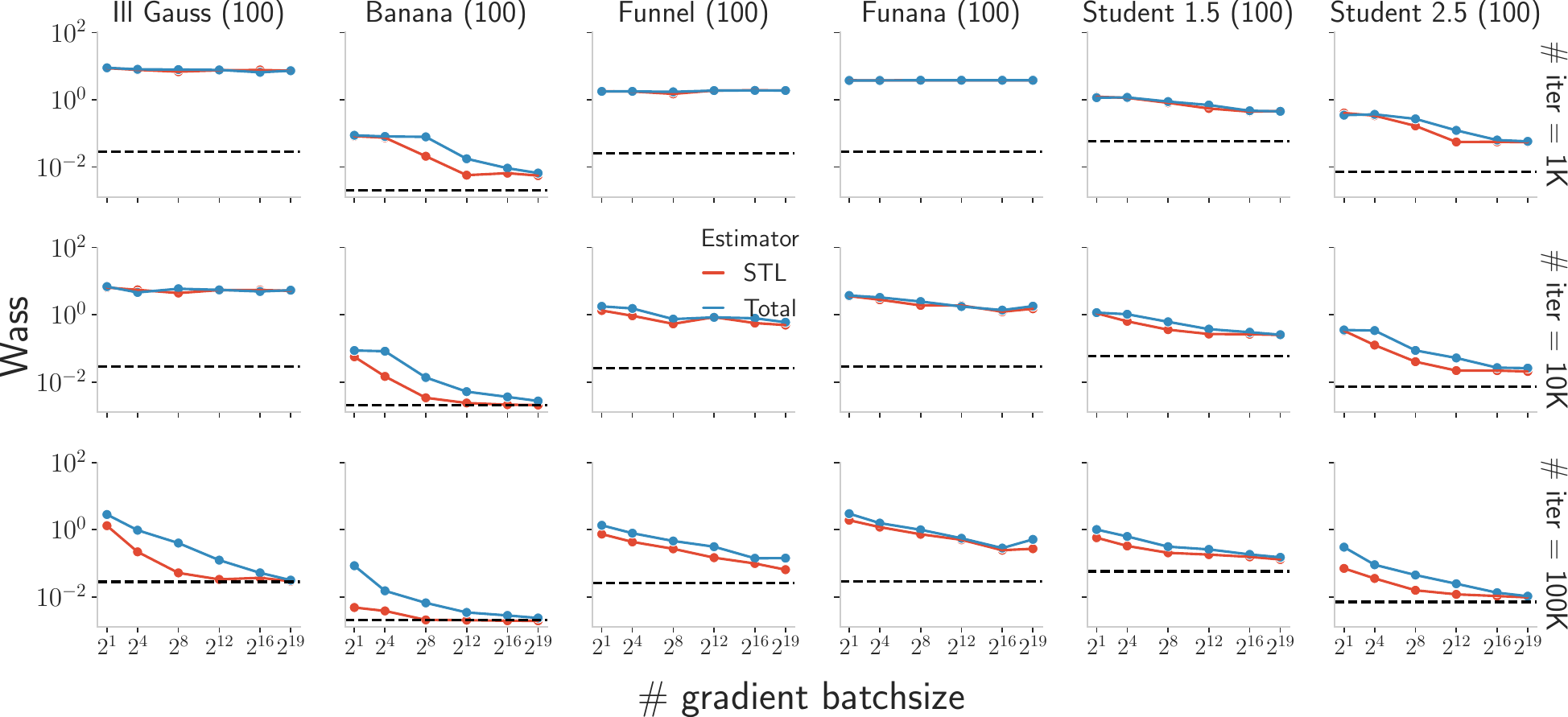}
    \captionsetup{justification=centering}
    % \caption{Results in 100 dimensions}
  \end{subfigure}
\caption{\small{\label{app: fig: estimators} \textbf{Rows:} Number of iterations. Dimensions indicated in brackets alongside model names. Figure uses the same setting as \Cref{fig: estimators} and includes the results for different dimensions (\Cref{fig: estimators} uses ten dimensions). }}
\end{figure}
We use a ten-layered real-NVP flow with $32$ hidden units and optimize $\KL{q}{p}$.
We run independent optimization for $1$K, $10$K and $100$K iterations using Adam, and sweep over three step-sizes: $0.0001, 0.0003,$ and$ 0.001.$ 
\Cref{fig: estimators} uses the decayed step-schedule. \Cref{app: fig: estimators} plots the results for different dimensions.

% \newpage
% We use a gradient batchsize of $512\times2^{10}$ samples, that i\s, at each iteration, we draw a fresh batch of $512\times 2^{10}$ samples from $q$ to approximate the $\KL{q}{p}$ gradients. 

\section{DETAILS FOR \Cref{sec:optimization} EXPERIMENTS}
\label{app: sec: details of optimization}
\begin{figure}[ht!]
  \centering
  \begin{subfigure}{\linewidth}
    % \begin{center}
      \centering
      \includegraphics[width = 0.8\textwidth]{./figures/section_6/scaled_wass_decay_schedule_new.pdf}
      \captionsetup{justification=centering} 
      \caption{Uses decay step-schedule.}
    % \end{center}
  \end{subfigure}
  \begin{subfigure}{\linewidth}
    \centering
    \includegraphics[width = 0.8\textwidth]{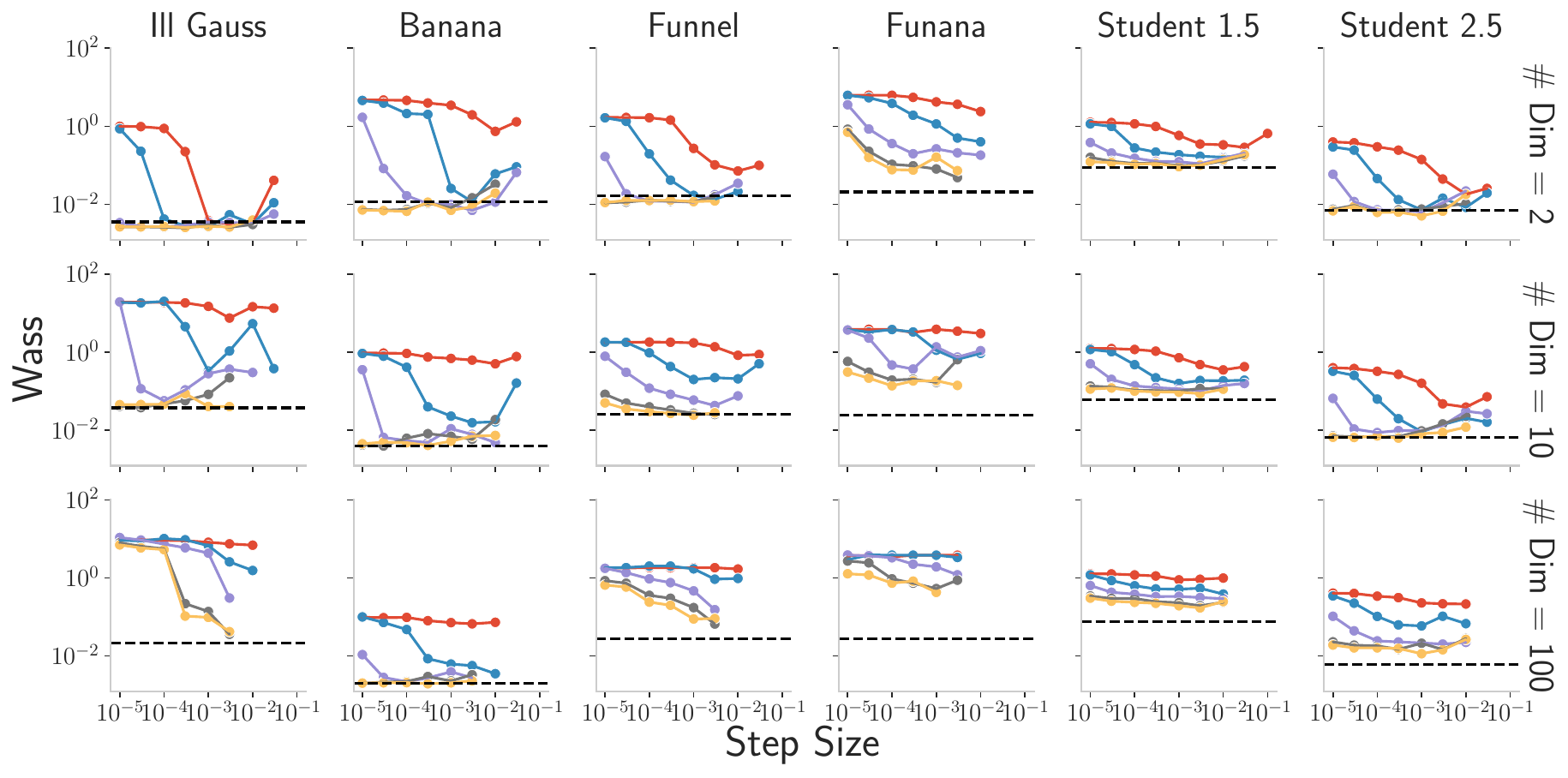}
    \captionsetup{justification=centering}
    \caption{Uses constant step-schedule.}
  \end{subfigure}
  % \caption{\small{\label{fig: optimization} \textbf{Rows:} Model dimensions. {\Margwass} metric against step-sizes for different number of iterations. Each point in the plot corresponds to an independent optimization run with missing values indicating divergence. Interestingly, for some step-sizes like $10^{-2}$ optimization diverges after exceptionally good performance for several thousand iterations (see performance for Funnel and Funana). Generally, step-sizes in the range of $[10^{-4}, 10^{-3}]$ work consistently across the models when optimized for $10$K iterations or more.}}
  \caption{
    \small{\label{app: fig: optimization} \textbf{Rows:} Model dimensions. Figure uses the same setting as \Cref{fig: optimization} and includes the results for different step-schedules (\Cref{fig: optimization} uses decay step-schedule). See \Cref{app: sec: details of capacity} for details about the decayed step-schedule.   }
  }
\end{figure}

We use a ten-layered real-NVP flow with $32$ hidden units and optimize $\KL{q}{p}$ using STL.  
We use a gradient batchsize of $512\times2^{10}$ samples, that is, at each iteration, we draw a fresh batch of $512\times 2^{10}$ samples from $q$ to approximate the $\KL{q}{p}$ gradients. 
We run independent optimization for $100$, $1$K, $10K$, $50$K, and $100$K iterations using Adam. \Cref{fig: optimization} in the paper uses the decayed step-size schedule. \Cref{app: fig: optimization} plots the results for different step-schedules.

% \newpage
\section{DETAILS FOR \Cref{sec:batch_size} EXPERIMENTS}
\label{app: sec: details of batch_size}
\begin{figure}[ht!]
  \centering
  \begin{subfigure}{\linewidth}
    \centering
    \includegraphics[width = 0.8\textwidth]{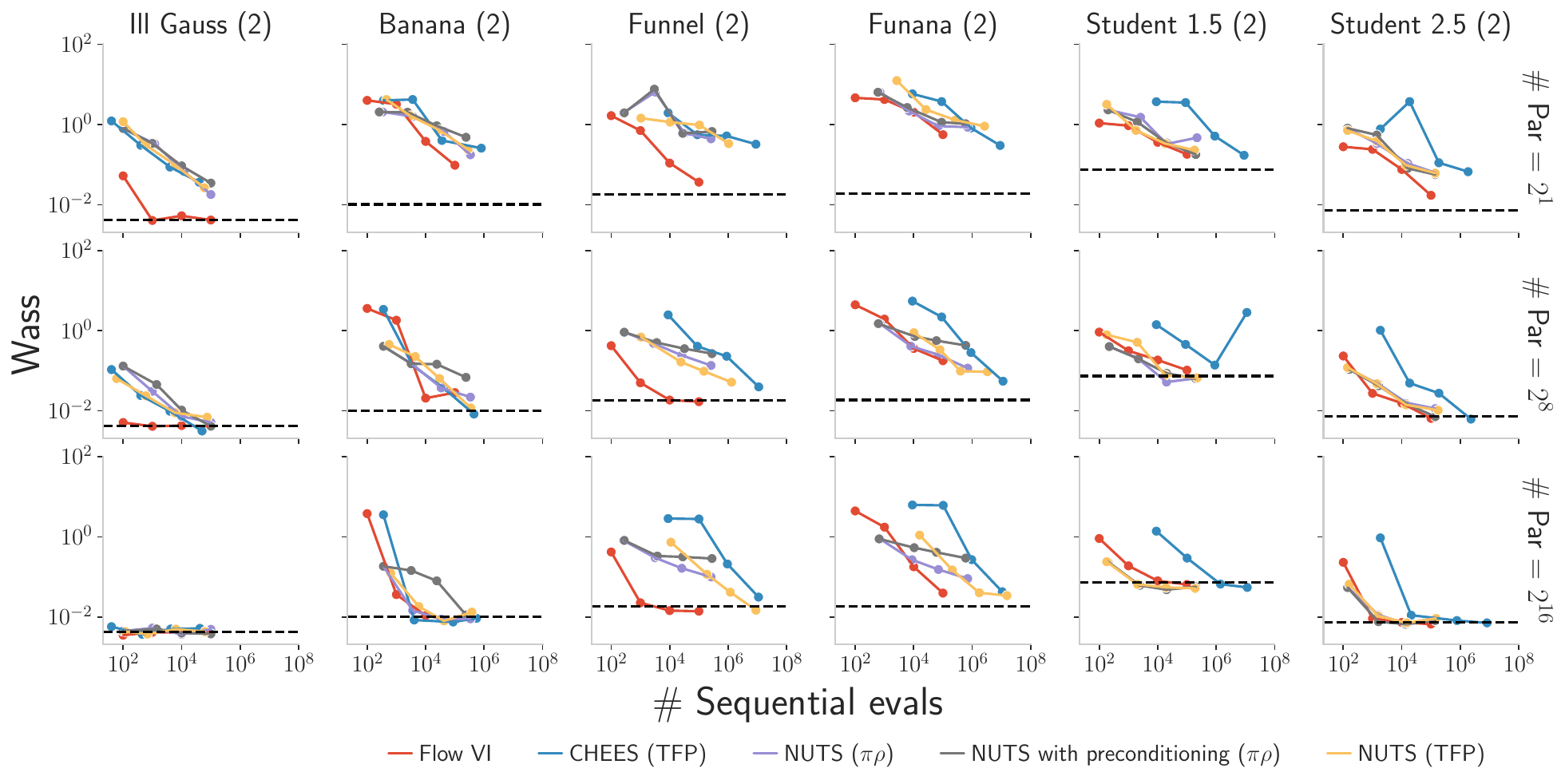}
    \captionsetup{justification=centering}
  \end{subfigure}
  \begin{subfigure}{\linewidth}
    \centering
    \includegraphics[width = 0.8\textwidth]{./figures/section_7/new_final_scaled_wass_values_all_methods_all_models_ndim_10_batchsizes____2__1___,___2__8___,___2__16_____all_hmc.pdf}
    \captionsetup{justification=centering}
  \end{subfigure}
  \begin{subfigure}{\linewidth}
    \centering
    \includegraphics[width = 0.8\textwidth]{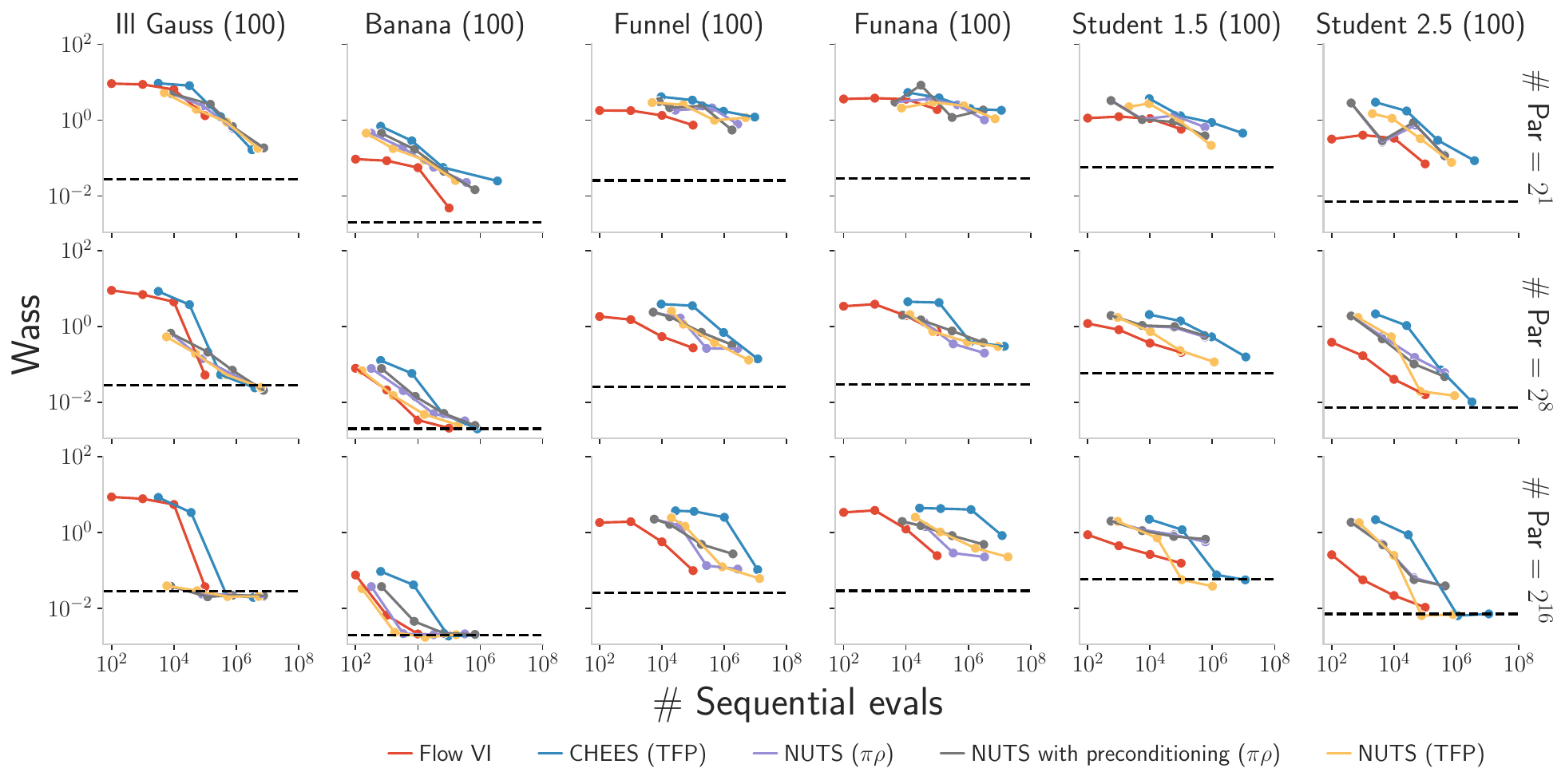}
    \captionsetup{justification=centering}
  \end{subfigure}
  \caption{\small{\label{app: fig: batchsize and hmc} \textbf{Rows:} Parallel evaluations. Dimensions indicated in brackets alongside model names. Figure uses the same setting as \Cref{fig: batchsize and hmc} and includes the results for different dimensions (\Cref{fig: batchsize and hmc} uses ten dimensions). }}
\end{figure}

\begin{figure*}[ht!]
  \centering
  \includegraphics[width = 0.8\textwidth, trim={0, 1.65cm, 0, 0}, clip]{./figures/section_7/real_models_without_correction__2,_64,_4096__all_models.pdf}
  \includegraphics[width = 0.8\textwidth, trim={0, 1.65cm, 0, 0}, clip]{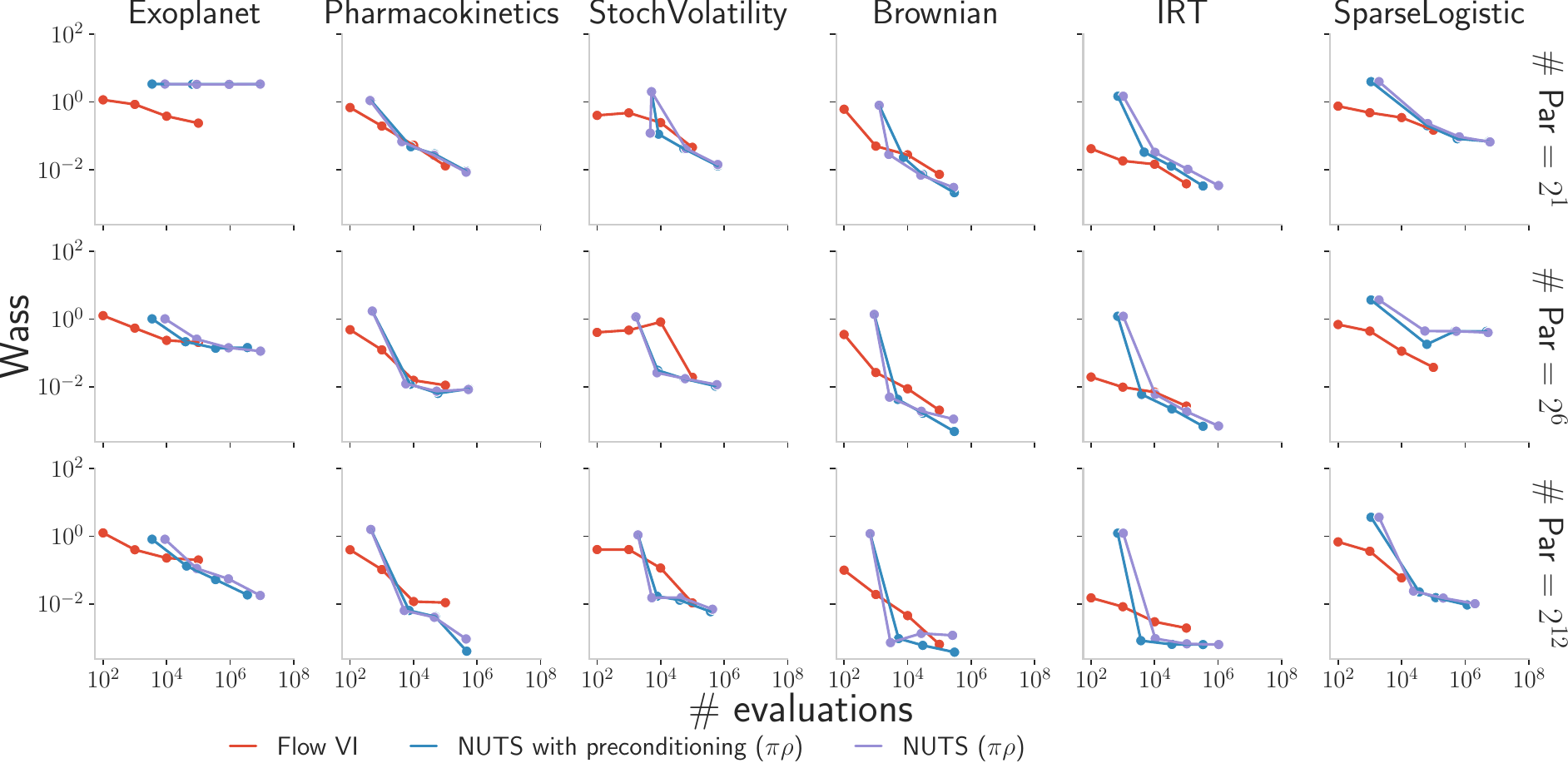}
  \includegraphics[width = 0.65\textwidth, trim={4cm, 1cm, 4cm, 14.5cm}, clip]{./figures/section_7/new_final_scaled_wass_values_all_methods_all_models_ndim_10_batchsizes____2__1___,___2__8___,___2__16_____all_hmc.pdf}\\
  \includegraphics[width = 0.65\textwidth, trim={5.5cm, 0, 1.5cm, 15.75cm}, clip]{./figures/section_7/new_final_scaled_wass_values_all_methods_all_models_ndim_10_batchsizes____2__1___,___2__8___,___2__16_____all_hmc.pdf}\\
  \caption{\small{
  \label{app: fig: real models} \textbf{Rows:} Parallel evaluations. {\Margwass} metric against number of sequential evaluations for non-synthetic models from \Cref{app: sec: details on non-synthetic models} with parallel evaluations increasing across the rows. 
  The first subplot corresponds to Wasserstein values when we do not filter the collapsed chains, and the second subplot corresponds to Wasserstein values when we filter the collapsed chains. }}
\end{figure*}

For flow VI, we use a ten-layered real-NVP flow with $32$ hidden units, we optimize $\KL{q}{p}$ using STL estimator. 
We run independent optimization for $100$, $1$K, $10$K, and $100$K iterations using Adam and use a single step-size of $3\times 10^{-4}$ along with a decayed step-schedule for $100$K iterations and a constant step-size of $3\times 10^{-4}$ for the rest.

For NUTS as implemented in TensforFlow Probability, we use a target acceptance probability of $0.75$, a starting step-size of $0.1$, and a maximum-tree-depth of $2^{10}$ steps. For CHEES as implemented in TensforFlow Probability, we use a target acceptance probability of $0.75$ and an initial step-size of $1.0$.

For NUTS as implemented in NumPyro, we use a target acceptance probability of $0.9$, a starting step-size of $0.1$, and a maximum-tree-depth of $2^{10}$. 
We experiment with preconditioning by turning the $\mathtt{adapt\_mass\_matrix}$ flag on and off, giving us two variants from NumPyro. 

For all HMC methods, we run chains for $10$, $100$, $1$K, and $10$K iterations and use as many warm-up iterations, where each iteration can include several leapfrog steps. Currently, most frameworks do not allow tracking the number of leapfrog steps during the warm-up phase. So, we extrapolate the number of leapfrog steps from post-warm-up leapfrog steps of the run with $256$ chains and $1$K iterations for the synthetic targets and $1024$ chains and $1$K iterations for the non-synthetic models. 

For all methods, we experiment with parallel compute budgets. 
For synthetic targets, we use a budget of $2, 2^{8},$ and $ 2^{16}$. 
For VI, this means that we use as many samples for evaluating the gradient, and for HMC, this means that we run as many chains in parallel. 
HMC chains either collect fewer than or more than a million samples, but not exactly a million samples when run for as many iterations as indicated above (we use a million samples for evaluations). 
For the chains that collect fewer samples, we simply use as many samples as possible from the chain. 
For the chains that collect more samples, we use thinning if necessary for memory constraints and use the last million samples collected. For instance, when using a budget of $2^{16}$ for the non-synthetic models, we run chains for $10$K iterations but collect only every $16$th sample, resulting in $1048576$ samples.
\Cref{app: fig: batchsize and hmc} plots the results across different dimensions.

For the non-synthetic models, we use a budget of $2$, $2^{6}$, and $2^{12}$. 
Again, we use similar procedure as above to get the last million samples from the chains that collect more than a million samples. 
\Cref{app: fig: real models} plots the results for different models.
On non-synthetic models, some of the chains get stuck in a region of low probability and do not collect useful samples. We make two plots in \Cref{app: fig: real models}. The first plot shows the Wasserstein distance when we do not filter the collapsed chains and the second plot shows the Wasserstein distance when we filter the collapsed chains. 
For details of the reference samples, see \Cref{app: sec: details on reference samples for non-synthetic models}. 

\subsection{Reference Samples for Non-Synthetic Models}
\label{app: sec: details on reference samples for non-synthetic models}
For non-synthetic models, we need reference samples to evaluate the marginal Wasserstein distance. 
For this, we run Numpyro's NUTS with preconditioning (as this performed the best based on preliminary experiments) for $2^{14}$ chains in parallel and $10$K iterations (we thin to collect a million samples in total). 
However, since we select model with non-trivial geometries (see \Cref{app: sec: details on non-synthetic models}), we found that several chains were stuck in a region of low probability (indicated by the sample being repeated for the entire duration of the run) and did not collect useful samples. Especially, for the Exoplanet model, the Pharmacokinetics model, and the Sparse Linear Regression model. (We run things in 32-bits. The performance may improve when using 64-bits.) For reference samples, we drop all such instances of stuck chains. This leaves us with at least $497395$ for all the models. So, we use $497395$ samples for reference samples for all non-synthetic models. When using reference samples, we use as many samples from a method as possible. So, if a method collects more than $497395$ samples, we use $497395$ samples. If a method collects less than $497395$ samples, we use all samples collected.

\section{TARGET DETAILS}
\label{app: sec: details on other models}
% \begin{itemize}[leftmargin=15pt,itemsep=0pt,topsep=0pt]
\subsection{Synthetic Targets}
\label{app: sec: details on synthetic models}
\textbf{Ill-conditioned Gaussian.} A multivariate Gaussian with zero mean and a covariance matrix with eigenvalues sampled from a gamma distribution (shape = 0.5, scale = 1) and then rotated by a random orthogonal matrix, ensuring high correlations. We use Inference Gym implementation to ensure reproducibile targets \citep{inferencegym2020}.

\textbf{Banana \citep{haario1999adaptive}.} 
This distribution modifies a 2-dimensional normal distribution by applying a non-linear transformation to the second dimension, introducing dependence on the first dimension. The transformation involves a curvature parameter that modifies the scale of the second dimension based on the value of the first dimension. If more than two dimensions are specified, the additional dimensions are modeled as standard normal variables.

For $ d$ dimensions, the distribution is
\begin{align*}
z_1 &\sim \mathcal{N}(0, 10^2), \\
z_2 &\sim \mathcal{N}(0.03 (z_1^2 - 100), 1), \text{ and} \\
z_i &\sim \mathcal{N}(0, 1) \quad \text{for } i = 3, \dots, d.
\end{align*}
where $ z_1 $ is the primary dimension, sampled from a normal distribution with a mean of 0 and a scale of 10.
$ z_2 $ is the transformed second dimension, where the mean is adjusted by $ 0.03 (z_1^2 - 100)$, incorporating a quadratic dependence on $ z_1 $.
$ z_i $ for $ i \geq 3 $ are independent and follow a standard normal distribution, i.e., $ \mathcal{N}(0, 1) $.

% Features a difficult geometry with a \emph{nonlinear} relationship. In two dimensions: $z_1 \sim \mathcal{N}(0, 10)$ and $z_2 \sim \mathcal{N}(0.03(z_1^2 - 100), 1)$. In higher dimensions, $z_n \sim \mathcal{N}(0, 1)$ for $n \geq 3$.

\textbf{Neal's funnel \citep{neal2001annealed,neal2003slice}.} 
This distribution is constructed by transforming a multi-dimensional Gaussian distribution, with an initial scale for the first dimension and an exponentially scaled variance for the remaining dimensions. The transformation creates a funnel shape, with a narrow “neck” that poses challenges for some sampling algorithms \citep{betancourt2017conceptual}, such as HMC. The funnel shape resembles the posterior distributions commonly found in centrally parameterized hierarchical models.
For $d$ dimensions, the distribution is:

\begin{align*}
z_1 &\sim \mathcal{N}(0, 3^2) \\
z_i &\sim \mathcal{N}(0, \exp(z_1)) \quad \text{for } i = 2, \dots, d
\end{align*}

where $z_1$ is the first dimension, with a normal distribution of mean 0 and scale $3$, setting the initial scale for the funnel.
$z_i$ for $i \geq 2$ are additional dimensions, with scales that depend on $z_1$ through the transformation $\exp(z_1 / 2)$. This transformation introduces increased variance as $z_1$ grows, widening the distribution as you move away from the origin.

% Exemplifies common issues in hierarchical models, where the dispersion of one variable depends on another. Known to be challenging for HMC-based samplers \citep{betancourt2017conceptual}. In two dimensions: $z_1 \sim \mathcal{N}(0, 3)$ and $z_2 \sim \mathcal{N}(0, \exp(z_1/2))$. In higher dimensions, $z_n \sim z_2$ for $n \geq 3$.

\textbf{Funana.}
This new distribution is a hybrid of Neal's Funnel and the Banana distribution, combining features that challenge sampling methods with both a funnel-like narrowing and a non-linear transformation of higher dimensions. The result is a distribution with complex dependencies and variable scaling that can pose significant difficulties for inference algorithms.

For $d \geq 3$ dimensions, the distribution is:

\begin{align*}
z_1 &\sim \mathcal{N}(0, 3^2) \\
z_2 &\sim \mathcal{N}(0, 10^2) \\
z_3 &\sim \mathcal{N}(0.03(z_1^2 - 100), \exp(z_1))
\end{align*}

For dimensions $n \geq 4$, we extend $z_3$ to all additional dimensions, such that,

\begin{align*}
z_n &\sim \mathcal{N}(0.03(z_1^2 - 100), \exp(z_1)) \quad \text{for } n \geq 4
\end{align*}

where $z_1$ sets the scale for the distribution, much like the leading variable in Neal's Funnel, introducing an exponentially increasing variance for the subsequent dimensions. 
$z_2$ follows a normal distribution with a larger scale, contributing to the initial shape of the distribution.
$z_3$ is influenced by both the curvature (as in the Banana distribution) and the scaling behavior (as in Neal's Funnel), making it dependent on $z_1$ for both mean and variance.
Additional dimensions $z_n$ for $n \geq 4$ inherit the complex transformation from $z_3$, extending the funnel and banana features across higher dimensions.

Funana is challenging to sample from due to the combined narrowing effect of the funnel shape and the curvature introduced by the banana transformation. This mixture of features creates non-linear dependencies and variable scaling, which can confound sampling algorithms.

% A new distribution combining the difficulties of Neal's Funnel and the Banana distribution. In three dimensions: $z_1 \sim \mathcal{N}(0, 3)$, $z_2 \sim \mathcal{N}(0, 10)$, and $z_3 \sim \mathcal{N}(0.03(z_1^2 - 100), \exp(z_1/2))$. In higher dimensions, $z_n \sim z_3$ for $n \geq 4$.

\textbf{Student-t.} 
A multivariate Student-t distribution centered at the origin with an identity scale matrix, controlled by the degrees of freedom parameter, \( \nu \). The distribution exhibits heavier tails as \( \nu \) decreases, making it useful for modeling data with outliers or extreme values. We consider two specific cases: \( \nu = 1.5 \) and \( \nu = 2.5 \).
% The distribution is defined as
% \begin{align*}
% z \sim \text{Student-t}(\nu, \mathbf{0}, \mathbf{I}).
% \end{align*}
For $ \nu = 1.5 $, the distribution has particularly heavy tails, with undefined covariance.
For $ \nu = 2.5 $, the distribution still has heavier tails than a normal distribution, though its covariance exists.
% 
% A multivariate Student-t distribution with location zero and identity scale, characterized by $\nu$ degrees of freedom. Smaller $\nu$ results in heavier tails. Two cases are considered: $\nu=1.5$ (where covariance does not exist) and $\nu=2.5$.

% \end{itemize}

\subsection{Non-synthetic Models}
\label{app: sec: details on non-synthetic models}
We also compare flow VI with HMC on models where the target density is not available in closed-form, and we do not have access to ground truth samples. The details of the models are as follows. 
\paragraph{Exoplanet ($\mathrm{Dim} = 7$).}
This model describes the dimming of a star's light during a transit event, where an exoplanet passes in front of its host star. By analyzing the star's light curve, this model infers properties of the exoplanet, such as its size and orbital period. The model includes both the transit shape and parameters characterizing the exoplanet.

The latent variables governing the transit and exoplanet properties are modeled with the following priors:
\begin{align*}
t_0 &\sim \mathcal{N}(2.26, 1^2), \\
P &\sim \mathrm{LogNormal}(3.66, 0.1^2), \\
D &\sim \mathrm{LogNormal}(0.5, 0.1^2), \\
r &\sim \mathrm{LogNormal}(0.08, 0.1^2), \\
\tilde{b} &\sim \text{Uniform}(0, 1), \quad b = \tilde{b} \times (1 + r), \\
u &\sim p(u),
\end{align*}
where $ t_0 $ is the reference time of the transit, assumed to be around $ 2.26 $ with a standard deviation of $ 1 $. $ P $ is the orbital period, with $ \log P $ centered at $ 3.66 $ and a standard deviation of $ 0.1 $.
$ D $ is the transit duration, with $ \log D $ centered at $ 0.5 $ and a standard deviation of $ 0.1 $.
$ r $ is the planet-to-star radius ratio, with $ \log r $ centered at $ 0.08 $ and a standard deviation of $ 0.1 $.
$ \tilde{b} $ is an unscaled impact parameter, uniformly distributed between $ 0 $ and $ 1 $, which is scaled by $ 1 + r $ to obtain $ b $.
$ u $ represents the quadratic limb-darkening parameters, distributed as per \citet{kipping2013efficient}, as implemented in jaxoplanet \citep{jaxoplanet}.
The set of latent variables is $ z = \{ t_0, P, D, r, b, u \} $.

The model predicts the flux $ y_i^{\text{pred}} $ at each observation time $ t_i $ as follows:

\begin{align*}
y_i^{\text{pred}} = L(t_i; z)
\end{align*}

where $ L(t_i; z) $ is the transit light curve model incorporating orbital mechanics and limb-darkening effects \citep{agol2020analytic}, as implemented in jaxoplanet \citep{jaxoplanet}.

The observed flux $ y_i $ at time $ t_i $ is modeled as a normal distribution around the predicted flux:

\begin{align*}
y_i \sim \mathcal{N}(y_i^{\text{pred}}, 0.03^2)
\end{align*}

where $ 0.03 $ is the observational noise standard deviation.

The joint distribution of the observed flux $ y $ and latent variables $ z $ is given by:

\begin{align*}
p(y, z) = \left( \prod_i \mathcal{N}(y_i \mid L(t_i; z), 0.03^2) \right) \times p(z)
\end{align*}

where $ p(z) $ represents the prior distributions of the latent variables as specified above.

Observations for $ y $ are generated using ancestral sampling over a time range $ t $ from $ 0 $ to $ 17 $ with a step size of $ 0.05 $. We condition on data generated from this process. We work in the unconstrained space for ease, so use log-transformations for the parameters $ P $, $ D $, and $ r $.

\paragraph{Pharmacokinetics ($\mathrm{Dim} = 45$).}

This hierarchical model describes the absorption and distribution of an orally administered drug in the body, using a one-compartment pharmacokinetic system with first-order absorption. The model captures both population-level parameters and individual variability across \( n = 20 \) patients, enabling it to reflect differences in drug absorption and elimination rates between individuals.

The drug is absorbed from the gut into the bloodstream and distributed to various organs. This process is governed by the following system of differential equations, representing the rate of change in drug concentration over time:
\begin{align*}
\frac{dm_{\text{gut}}}{dt} &= -k_1 m_{\text{gut}} \\
\frac{dm_{\text{cent}}}{dt} &= k_1 m_{\text{gut}} - k_2 m_{\text{cent}}
\end{align*}
where \( m_{\text{gut}}(t) \) is the mass of the drug in the gut compartment. \( m_{\text{cent}}(t) \) is the mass of the drug in the central compartment (e.g., bloodstream). \( k_1 \) is the rate constant for absorption from the gut to the central compartment. \( k_2 \) is the rate constant for elimination from the central compartment.

For \( k_1 \neq k_2 \), the system has an analytical solution that provides the drug concentration over time:

\begin{align*}
m_{\text{gut}}(t) &= m_{\text{gut}}^0 \exp(-k_1 t) \\
m_{\text{cent}}(t) &= \frac{\exp(-k_2 t)}{k_1 - k_2} \left( m_{\text{gut}}^0 k_1 \left(1 - \exp([k_2 - k_1] t)\right) + (k_1 - k_2) m_{\text{cent}}^0 \right)
\end{align*}
where \( m_{\text{gut}}^0 \) and \( m_{\text{cent}}^0 \) are the initial amounts of drug in the gut and central compartments, respectively.

Patients receive repeated doses at times \( 0 \), \( 12 \), and \( 24 \) hours. The model incorporates these dosing events by updating the amount of drug in the gut compartment at each dosing time, effectively resetting the boundary conditions for the differential equations. Each patient’s observations are taken at various time points after each dose, covering a total time range from \( 0 \) to \( 32 \) hours.

The model uses a hierarchical Bayesian framework to estimate patient-specific parameters \( k_1^n \) and \( k_2^n \), based on population-level parameters with non-centered parameterization
\begin{align*}
k_{1,\text{pop}} &\sim \text{LogNormal}(\log 1, 0.1), \\
k_{2,\text{pop}} &\sim \text{LogNormal}(\log 0.3, 0.1) \\
\sigma_{1} &\sim \text{LogNormal}(\log 0.15, 0.1), \\
\sigma_{2} &\sim \text{LogNormal}(\log 0.35, 0.1), \\
\sigma &\sim \text{LogNormal}(-1, 1)
\end{align*}

For each patient \( n \), the individual parameters are given by:
\begin{align*}
\eta_{1}^n \sim \mathcal{N}(0, 1), \quad \eta_{2}^n \sim \mathcal{N}(0, 1)
\end{align*}

\begin{align*}
k_{1}^n &= k_{1,\text{pop}} \cdot \exp(\eta_{1}^n \sigma_{1}), \\
k_{2}^n &= k_{2,\text{pop}} \cdot \exp(\eta_{2}^n \sigma_{2})
\end{align*}

The observed drug concentration \( y_n \) for patient \( n \) at a given time is modeled as:
\begin{align*}
y_n \sim \text{LogNormal}\left(\log m_{\text{cent}}(t; k_{1}^n, k_{2}^n), \sigma\right)
\end{align*}

where \( m_{\text{cent}}(t; k_{1}^n, k_{2}^n) \) is the concentration in the central compartment, computed using the analytical solution with patient-specific rate constants.

We condition on data generated from ancestral sampling. 
We work in the unconstrained space for ease, so use log-transformations for the parameters $k_1$, $k_2$, $\sigma_1$, $\sigma_2$, and $\sigma$.

        \paragraph{Stochastic Volatility ($\mathrm{Dim} = 103$).}
        This model captures the volatility of asset prices over time, assuming that volatility follows an AR(1) process with an unknown persistence coefficient and shock scale. The model uses vectorized computations for efficiency and assumes that daily returns are centered around zero with a time-varying volatility.
        For $ T $ timesteps, the model is:
        
        \begin{align*}
          \phi &\sim 2 \times \text{Beta}(20, 1.5) - 1 \\
          \mu &\sim \text{Cauchy}(0, 5) \\
          \sigma_w &\sim \text{HalfCauchy}(0, 2)
\end{align*}

The log volatility $ h_t $ follows an AR(1) process:
\begin{align*}
h_0 &\sim \mathcal{N}\left(0, \frac{\sigma_w}{\sqrt{1 - \phi^2}}\right) \\
h_t &\sim \mathcal{N}(\phi \, h_{t-1}, \sigma_w) \quad \text{for } t = 1, \dots, T-1
\end{align*}

The returns $ y_t $ are modeled with a time-varying volatility:

\begin{align*}
y_t &\sim \mathcal{N}\left(0, \sqrt{\exp(\mu + h_t)}\right) \quad \text{for } t = 0, \dots, T-1
\end{align*}

where $ \phi $ is the persistence of volatility, rescaled to lie between $[-1, 1]$ using a Beta distribution; $ \mu $ represents the mean log volatility; $ \sigma_w $ is the white noise shock scale, determining the variability of the AR(1) process driving the log volatility; $ h_t $ is the latent log volatility at time $ t $, evolving as an AR(1) process; $ y_t $ is the centered return at time $ t $, assumed to have a time-varying standard deviation based on the current log volatility.

There are three top-level parameters and 100 per-time-step latent variables, for a total of 103 dimensions. We condition on data $ y $ drawn from the prior.

\paragraph{Item Response Theory ($\mathrm{Dim} = 501$).}

This model describes the probability of a set of students answering a set of questions correctly, based on each student’s ability and each question’s difficulty. The model assumes a one-parameter logistic form, where each student has an individual ability parameter, and each question has a difficulty parameter. Additionally, there is a shared mean ability across all students.
For a group of students and questions, the model is:

\begin{align*}
\mu      &\sim \mathcal{N}(0.75, 1) \\
\alpha_i &\sim \mathcal{N}(0, 1) \quad \text{for } i = 1, \dots, N_{\text{students}} \\
\beta_j  &\sim \mathcal{N}(0, 1) \quad \text{for } j = 1, \dots, N_{\text{questions}}
\end{align*}

For each student-question pair $(i, j)$, the probability of a correct answer is modeled as:

\begin{align*}
y_{i,j} \sim \text{Bernoulli}(\sigma(\alpha_i - \beta_j + \mu))
\end{align*}

where $ \mu $ is the mean student ability, shared across all students; $ \alpha_i $ is the centered ability of student $ i $, reflecting their specific skill level; $ \beta_j $ is the difficulty of question $ j $, representing how challenging the question is; $ y_{i,j} $ is a binary indicator of whether student $ i $ answered question $ j $ correctly; $ \sigma(\cdot) $ is the logistic sigmoid function, translating the linear combination into a probability.

There are $N_{\text{students}} = 100 $ students, $ N_{\text{questions}} = 400 $ questions, and $ N_{\text{responses}} = 30105 $ responses, for a total of 501 parameters. We condition on data drawn from the prior.

\paragraph{Sparse Logistic Regression ($\mathrm{Dim} = 51$).}

This is a hierarchical logistic regression model with a sparse prior applied to the German credit dataset. 
We use the variant of the dataset, with the covariates standardized to range between -1 and 1. With the addition of a constant factor, this yields 25 covariates.
The model is defined as follows:

\begin{align*}
\tau &\sim \text{Gam}(\alpha = 0.5, \beta = 0.5) \\
\lambda_d &\sim \text{Gam}(\alpha = 0.5, \beta = 0.5) \\
\beta_d &\sim \mathcal{N}(0, 1) \quad (10)\\
y_n &\sim \text{Bern}(\mathrm{sigmoid}(x_n^\top (\tau \beta \circ \lambda)))
\end{align*}

where Gam is the Gamma distribution, $ \tau $ is the overall scale, $ \lambda $ are per-dimension scales, $ \beta $ are the non-centered covariate weights, $ \beta \circ \lambda $ denotes the elementwise product of $ \beta $ and $ \lambda $, and $ \mathrm{sigmoid} $ is the sigmoid function. The sparse gamma prior on $ \lambda $ imposes a soft sparsity prior on the weights, which could be used for variable selection. This parameterization uses $ D = 51 $ dimensions. We log-transform $ \tau $ and $ \beta $ to make them unconstrained.

\paragraph{Brownian Motion ($\mathrm{Dim} = 32$).}
This model represents a Brownian Motion process with Gaussian observations, where the scale parameters for both the process noise and observation noise are unknown and modeled with log-normal distributions.
For $ T $ timesteps, the model is:

  \begin{align*}
\sigma_{\text{proc}} &\sim \text{LogNormal}(0, 2) \\
\sigma_{\text{obs}} &\sim \text{LogNormal}(0, 2) \\
x_0 &\sim \mathcal{N}(0, \sigma_{\text{proc}}) \\
x_t &\sim \mathcal{N}(x_{t-1}, \sigma_{\text{proc}}) \quad \text{for } t = 1, \dots, T-1 \\
y_t &\sim \mathcal{N}(x_t, \sigma_{\text{obs}}) \quad \text{for } t = 0, \dots, T-1
\end{align*}

where $ \sigma_{\text{proc}} $ is the process noise scale, which controls the variability in the Brownian Motion from one step to the next;
 $ \sigma_{\text{obs}} $ is the observation noise scale, determining the level of noise in the observed positions; and $ x_t $ represents the latent position at time $ t $, while $ y_t $ is the observed position subject to observation noise.
 In this model, we condition on data generated from ancestral sampling. We set the number of timesteps to $ T = 30 $ and fix the middle ten timesteps to NaNs to simulate missing values.

\end{document}